\documentclass[11pt]{article}
\usepackage[margin=1in]{geometry}
\usepackage[T1]{fontenc}
\usepackage{times}
\usepackage{microtype}
\usepackage{amsthm}

\newtheorem{theorem}{Theorem}[section]

\newtheorem{lemma}[theorem]{Lemma}

\theoremstyle{remark}

\theoremstyle{plain}

\newif\ifspacehack
\usepackage{natbib}
\usepackage{hyperref}
\usepackage{url} 
\usepackage{graphicx}
\usepackage{mathtools}
\usepackage{footnote}
\usepackage{float}
\usepackage{xspace}
\usepackage{multirow}
\usepackage{xcolor}
\usepackage{wrapfig}
\usepackage{framed}
\usepackage{bbm}
\usepackage{footnote}
\usepackage{nicefrac}
\usepackage{makecell}
\usepackage[algo2e,ruled,vlined]{algorithm2e} 
\usepackage{algorithm}
\usepackage{amssymb}
\usepackage{bm}
\makesavenoteenv{tabular}
\makesavenoteenv{table}

\def \R {\mathbb{R}}

\newcommand{\calA}{{\mathcal{A}}}

\newcommand{\calX}{{\mathcal{X}}}

\newcommand{\calF}{{\mathcal{F}}}

\newcommand{\calE}{{\mathcal{E}}}
\newcommand{\calG}{{\mathcal{G}}}

\newcommand{\calT}{{\mathcal{T}}}
\newcommand{\calP}{{\mathcal{P}}}

\newcommand{\Reg}{\mathbf{\mathbf{Reg}}}

\newcommand{\E}{{\mathbb{E}}}

\newcommand{\inner}[1]{ \left\langle {#1} \right\rangle }

\newcommand{\order}{\mathcal{O}}

\newcommand{\wh}{\widehat}

\newcommand{\otil}{\ensuremath{\tilde{\mathcal{O}}}}

\usepackage{lipsum,booktabs}
\usepackage{amsmath,amssymb,amsfonts,bm,enumitem}
\usepackage{rotating}
\usepackage{pdflscape}
\usepackage{hyperref,url}
\hypersetup{
    colorlinks,
    breaklinks,
    linkcolor = blue,
    citecolor = blue,
    urlcolor  = blue,
}
\allowdisplaybreaks
\usepackage{appendix}
\usepackage{multirow,makecell}

\usepackage{algorithmic,algorithm}

\renewcommand{\tilde}{\widetilde}
\renewcommand{\hat}{\widehat}

\def \E {\mathbb{E}}

\def \R {\mathbb{R}}

\def \Pr {\mathsf{Pr}}

\usepackage{mathtools}

\usepackage{graphicx,color}
\definecolor{wine_red}{RGB}{228,48,64}
\definecolor{DSgray}{cmyk}{0,1,0,0}

\usepackage{prettyref}
\newcommand{\pref}[1]{\prettyref{#1}}

\newcommand{\plines}[2]{\savehyperref{#1}{Lines~\ref*{#1}} and~\savehyperref{#2}{\ref*{#2}}}
\newcommand{\savehyperref}[2]{\texorpdfstring{\hyperref[#1]{#2}}{#2}}
\newrefformat{eq}{\savehyperref{#1}{Eq. \textup{(\ref*{#1})}}}
\newrefformat{eqn}{\savehyperref{#1}{Eq.~(\ref*{#1})}}
\newrefformat{lem}{\savehyperref{#1}{Lemma~\ref*{#1}}}
\newrefformat{def}{\savehyperref{#1}{Definition~\ref*{#1}}}
\newrefformat{line}{\savehyperref{#1}{Line~\ref*{#1}}}
\newrefformat{thm}{\savehyperref{#1}{Theorem~\ref*{#1}}}
\newrefformat{corr}{\savehyperref{#1}{Corollary~\ref*{#1}}}
\newrefformat{cor}{\savehyperref{#1}{Corollary~\ref*{#1}}}
\newrefformat{sec}{\savehyperref{#1}{Section~\ref*{#1}}}
\newrefformat{app}{\savehyperref{#1}{Appendix~\ref*{#1}}}
\newrefformat{assum}{\savehyperref{#1}{Assumption~\ref*{#1}}}
\newrefformat{asm}{\savehyperref{#1}{Assumption~\ref*{#1}}}
\newrefformat{ex}{\savehyperref{#1}{Example~\ref*{#1}}}
\newrefformat{fig}{\savehyperref{#1}{Figure~\ref*{#1}}}
\newrefformat{alg}{\savehyperref{#1}{Algorithm~\ref*{#1}}}
\newrefformat{rem}{\savehyperref{#1}{Remark~\ref*{#1}}}
\newrefformat{conj}{\savehyperref{#1}{Conjecture~\ref*{#1}}}
\newrefformat{prop}{\savehyperref{#1}{Proposition~\ref*{#1}}}
\newrefformat{proto}{\savehyperref{#1}{Protocol~\ref*{#1}}}
\newrefformat{prob}{\savehyperref{#1}{Problem~\ref*{#1}}}
\newrefformat{claim}{\savehyperref{#1}{Claim~\ref*{#1}}}
\newrefformat{que}{\savehyperref{#1}{Question~\ref*{#1}}}
\newrefformat{op}{\savehyperref{#1}{Open Problem~\ref*{#1}}}
\newrefformat{fn}{\savehyperref{#1}{Footnote~\ref*{#1}}}

\def \epsilon {\varepsilon}

\newcounter{restatesave}
\newcommand{\restatelemma}[2]{\setcounter{restatesave}{\value{theorem}}{\renewcommand{\thetheorem}{\ref*{#1}}\def\theHtheorem{restate.#1}#2}\setcounter{theorem}{\value{restatesave}}}

\RestyleAlgo{ruled}
\SetAlgoVlined
\DontPrintSemicolon
\newcommand{\indic}[1]{\mathbbm{1}\{#1\}}
\newcommand{\hsq}{\mathbb{H}^2}
\newcommand{\TV}{\mathrm{TV}}
\newcommand{\Pt}{\widetilde P}
\newcommand{\qt}{\widetilde q}
\newcommand{\qh}{\widehat q}
\newcommand{\Ph}{\widehat P}
\newcommand{\Pb}{\bar P}
\newcommand{\Vh}{\widehat V}
\newcommand{\ellh}{\widehat \ell}
\newcommand{\ellbar}{\bar \ell}
\newcommand{\Dpsi}{D_{\Psi}}
\newcommand{\dhat}{\widehat d}

\providecommand{\OccErr}{\textnormal{\textsc{Occ-Error}}}
\providecommand{\LrnBias}{\textnormal{\textsc{Learner-Bias}}}
\providecommand{\OmdReg}{\textnormal{\textsc{OMD-Regret}}}
\providecommand{\CompBias}{\textnormal{\textsc{Comp-Bias}}}
\providecommand{\ShiftPrice}{\textnormal{\textsc{Shift-Price}}}

\providecommand{\BOMD}{B$^2$OMD}

\usepackage{thmtools,thm-restate}

\title{\BOMD{}: Near-Optimal Regret Guarantee for Adversarial MDPs with Unknown Transitions and Bandit Feedback}
\author{  Mengxiao Zhang \\
  University of Iowa \\  \texttt{mengxiao-zhang@uiowa.edu} \\}
\date{}

\begin{document}
\maketitle

\begin{abstract}
We study episodic adversarial Markov decision processes with unknown transitions and bandit feedback. The learner interacts for $T$ episodes with a layered MDP of horizon $L$, a total of $X$ states across all layers, and $A$ actions at each decision state. Several works have studied this problem. However, a gap remains between their regret bounds and the $\Omega(L\sqrt{XAT})$ lower bound~\citep{jin2018q,domingues2021episodic}. In this paper, we propose Bias-Injected Block Occupancy Mirror Descent (\BOMD{}), an efficient algorithm that achieves $\otil(L\sqrt{XAT})$ expected regret. To our knowledge, this is the first efficient algorithm that matches the lower bound up to logarithmic factors. Our algorithm builds on UOB-REPS proposed by~\citet{jin2020learning} with several additional technical components, including a bias injected into the loss estimate through a value estimate, a block-wise update that holds the policy fixed over several episodes, and a hybrid regularizer that controls the stability of the resulting mirror descent update.
\end{abstract}

\section{Introduction}\label{sec:intro}

In this paper, we study episodic adversarial Markov decision processes (MDPs) with unknown transitions and bandit feedback, a model that extends adversarial multi-armed bandits to sequential decision making \citep{evendar2009online,neu2010online,zimin2013online,rosenberg2019online,jin2020learning}. In each episode, the learner chooses a policy and moves through a layered MDP whose transition is fixed but unknown, while the losses of the state-action pairs may change arbitrarily from episode to episode. The learner observes only the losses of the state-action pairs it visits, and its objective is to minimize the regret, that is, the difference between its expected cumulative loss and that of the best fixed policy in hindsight.

When the transition is known, \citet{zimin2013online} proposed O-REPS, which achieves $\otil(\sqrt{LXAT})$ regret under bandit feedback, where $L$ denotes the horizon, $X$ denotes the number of states, $A$ denotes the number of actions at each state, and $T$ denotes the number of episodes.\footnote{Throughout the paper, $\otil(\cdot)$ hides polylogarithmic factors in $L$, $X$, $A$, and $T$.} \citet{zimin2013online} also provide a matching $\Omega(\sqrt{LXAT})$ lower bound, showing that the obtained upper bound is optimal up to logarithmic factors. However, when the transition is unknown, the minimax regret is not yet settled. \citet{rosenberg2019bandit} first proposed an algorithm achieving $\otil(L^{3/2}XA^{1/4}T^{3/4})$ regret, while \citet{jin2020learning} then proposed an algorithm called UOB-REPS, which achieves $\otil(LX\sqrt{AT})$ regret. Subsequent works obtained the same $\otil(LX\sqrt{AT})$ regret with various approaches \citep{lee2020bias,dai2022follow,li2026closing}, and, very recently, \citet{kulkarni2026introducing} obtained $\otil(L^{5/2}\sqrt{XAT})$ regret by introducing feedback delays. On the other hand, the best known lower bound is $\Omega(L\sqrt{XAT})$ \citep{jin2018q,domingues2021episodic}, which is still a factor of $\sqrt{X}$ away from the bound of UOB-REPS and a factor of $L^{3/2}$ away from that of \citet{kulkarni2026introducing}. This leads to the following open question \citep{jin2020learning,li2026closing}:
\begin{center}
\emph{What is the minimax regret of adversarial MDPs with unknown transitions and bandit feedback, and how can we design an efficient algorithm that achieves it?}
\end{center}

\paragraph{Contributions.}
In this paper, we resolve this open question by proposing Bias-Injected Block Occupancy Mirror Descent (\BOMD{}), an efficient algorithm that achieves $\otil(L\sqrt{XAT})$ expected regret. This matches the $\Omega(L\sqrt{XAT})$ lower bound \citep{jin2018q,domingues2021episodic} up to logarithmic factors and additive lower-order terms, which shows that the minimax regret is $\widetilde{\Theta}(L\sqrt{XAT})$ when $T$ is sufficiently large. To our knowledge, this is the first efficient algorithm that matches the lower bound in this setting. Notably, the $\Omega(L\sqrt{XAT})$ lower bound holds even under full-information feedback, since its hard instances share a fixed loss function and differ only in the transition \citep{domingues2021episodic}. As full-information feedback is easier than bandit feedback, our result also achieves the minimax regret under full-information feedback with unknown transitions, showing that observing the losses of all state-action pairs does not help up to logarithmic factors.

Technically, \BOMD{} builds on UOB-REPS \citep{jin2020learning}, which runs mirror descent over the occupancy measures of the transitions in a confidence set with an optimistic loss estimate, with several important modifications. Specifically, we introduce (i) a bias injected into the loss estimate through a value estimate under the lower endpoints of the confidence intervals, together with a correction factor on the loss estimate, so that the errors caused by the unknown transition are bounded by the squared Hellinger distance between transitions in the confidence set; (ii) a block-wise update that holds the policy fixed over a block of episodes, so that the two parts of the bias, as well as the loss estimate and its correction factor, can be computed from separate halves of the block and are thus independent of each other; and (iii) a hybrid regularizer that combines the negative entropy with a log-barrier, which controls the stability of the resulting mirror descent update.

\subsection{Related Work}\label{sec:related}

\paragraph{Adversarial MDPs with full-information feedback.}
Online MDPs with adversarial losses were introduced by \citet{evendar2009online}, and the episodic loop-free formulation goes back to \citet{neu2010online}. With a known transition, O-REPS achieves $\otil(L\sqrt{T})$ regret under full-information feedback \citep{zimin2013online}. With an unknown transition, \citet{neu2012adversarial} first study this setting, and \citet{rosenberg2019online} propose an online learning algorithm over the occupancy measures of a confidence set that achieves $\otil(LX\sqrt{AT})$ regret. More recently, \citet{tiapkin2025narrowing} improve the dependence on $X$ to $\otil(L^{3}\sqrt{XAT})$ with policy optimization. Very recently, \citet{schlisselberg2026improved} further improve this bound to $\otil(L^{3/2}\sqrt{XAT})$ with variance-aware policy optimization, which is still a factor of $\sqrt{L}$ away from the $\Omega(L\sqrt{XAT})$ lower bound \citep{jin2018q,domingues2021episodic}. This lower bound holds even under full-information feedback because its hard instances differ only in the transition, and our bound matches it in this setting as well, since any algorithm for bandit feedback also applies under full-information feedback.

\paragraph{Adversarial MDPs with bandit feedback.} With a known transition, O-REPS achieves the optimal $\otil(\sqrt{LXAT})$ regret with bandit feedback \citep{zimin2013online}. With an unknown transition, \citet{rosenberg2019bandit} first obtain $\otil(L^{3/2}XA^{1/4}T^{3/4})$ regret, and \citet{jin2020learning} propose UOB-REPS, the first algorithm with $\otil(LX\sqrt{AT})$ regret. Building on occupancy-measure methods, subsequent works establish various types of guarantees under unknown transitions, including high-probability data-dependent bounds \citep{lee2020bias}, best-of-both-worlds guarantees \citep{jin2021best}, follow-the-perturbed-leader analogues \citep{dai2022follow}, and bounds under delayed feedback \citep{lancewicki2022learning,jin2022near}, all with at least the same $LX\sqrt{AT}$ dependence. Another line of work studies policy optimization, which updates the policy locally at each state in closed form and avoids optimization over the occupancy-measure polytope. \citet{shani2020optimistic} obtain the first sublinear regret of $\otil(T^{2/3})$ with policy optimization. Subsequently, \citet{luo2021policy} improve it to $\otil(L^2X\sqrt{AT})$ with dilated bonuses. \citet{dann2023best} further show that policy optimization can achieve best-of-both-worlds guarantees, and \citet{li2026closing} recently close the remaining horizon gap and obtain $\otil(LX\sqrt{AT})$ regret. Very recently, \citet{kulkarni2026introducing} obtain $\otil(L^{5/2}\sqrt{XAT})$ regret by introducing feedback delays. Our algorithm follows the occupancy-measure line and improves both the $\sqrt{X}$ factor of UOB-REPS and the $L^{3/2}$ factor of \citet{kulkarni2026introducing}.

\section{Preliminaries}\label{sec:prelim}

\paragraph{Problem setting.}
Following the formulation in prior work \citep{neu2012adversarial,zimin2013online,rosenberg2019online,jin2020learning}, we consider an episodic loop-free MDP with a layered state space $\calX=\calX_0\cup \calX_1\cup\cdots\cup \calX_L$, where the layers are disjoint and nonempty, $\calX_0=\{x_0\}$, and $\calX_L=\{x_L\}$. For a state $x\in \calX$, $h(x)\in\{0,\ldots,L\}$ denotes its layer, that is, $x\in \calX_{h(x)}$. The action set $\calA$ is finite. We write $X\triangleq|\calX|$ and $A\triangleq|\calA|$ for the numbers of states and actions, so that $L<X$ since every layer is nonempty. For each state $x\in \calX\setminus\{x_L\}$ and action $a\in \calA$, the unknown transition $P(\cdot|x,a)$ is a distribution over the next layer $\calX_{h(x)+1}$. A policy $\pi$ assigns to each state $x\in \calX\setminus\{x_L\}$ a distribution $\pi(\cdot|x)$ over $\calA$, and $\Pi$ denotes the set of all policies. Before the interaction, an oblivious adversary fixes losses $\ell_1,\ldots,\ell_T\in[0,1]^{\calX\times \calA}$. In episode $t\in\{1,\ldots,T\}$, the learner picks a policy $\pi_t\in\Pi$ and starts at $x_0$. It then generates the trajectory $(x_{t,0},a_{t,0},x_{t,1},\ldots,x_{t,L-1},a_{t,L-1},x_{t,L})$, where $x_{t,0}=x_0$, $a_{t,h}\sim\pi_t(\cdot|x_{t,h})$, and $x_{t,h+1}\sim P(\cdot|x_{t,h},a_{t,h})$ for $h\in\{0,\ldots,L-1\}$, and it observes the trajectory and the losses $\ell_t(x_{t,h},a_{t,h})$. We say that a state $x$ is visited in episode $t$ if $x_{t,h(x)}=x$, and that a pair $(x,a)$ is visited in episode $t$ if moreover $a_{t,h(x)}=a$.

\paragraph{Functions and sub-transitions.}
We use vector inequalities entrywise. In particular, $f\ge0$ means that every entry of $f$ is nonnegative, and $f>0$ means that every entry is positive. We call a vector in $\R^\calX$ a function on states and a vector in $\R^{\calX\times \calA}$ a function on pairs. Since no action is taken at $x_L$, the entries of a function on pairs at $x_L$ play no role, and for two functions $f,g$ on pairs, we write $\inner{f,g}\triangleq\sum_{x\in \calX\setminus\{x_L\}}\sum_{a\in \calA}f(x,a)g(x,a)$. A function $\varphi$ on states is also viewed as the function $(x,a)\mapsto\varphi(x)$ on pairs. More generally than a transition, a \emph{sub-transition} $\Pt$ assigns to each pair $(x,a)$ with $x\ne x_L$ nonnegative weights $\Pt(x'|x,a)$ on the states $x'\in \calX_{h(x)+1}$ with $\sum_{x'\in \calX_{h(x)+1}}\Pt(x'|x,a)\le1$. For two sub-transitions $\Pt$ and $\Pt'$, we write $\Pt\le\Pt'$ if $\Pt(x'|x,a)\le\Pt'(x'|x,a)$ for all $x\in \calX\setminus\{x_L\}$, $a\in \calA$, and $x'\in \calX_{h(x)+1}$. For a sub-transition $\Pt$ and a function $\varphi$ on states, $\Pt\varphi$ is the function on pairs defined by
$$[\Pt\varphi](x,a)\triangleq\sum_{x'\in \calX_{h(x)+1}}\Pt(x'|x,a)\varphi(x'),$$
so that $\Pt\varphi-\varphi$ is the function $(x,a)\mapsto[\Pt\varphi](x,a)-\varphi(x)$ on pairs.

\paragraph{Occupancy measures, values, and regret.}
For a sub-transition $\Pt$ and a policy $\pi$, the occupancy measure $q^{\Pt,\pi}$ is defined forward from $q^{\Pt,\pi}(x_0)\triangleq1$ by setting, for every state $x\in\calX\setminus\{x_L\}$, every action $a\in\calA$, and every state $x'\ne x_0$,
\begin{align*}
q^{\Pt,\pi}(x,a)&\triangleq q^{\Pt,\pi}(x)\,\pi(a|x),\\
q^{\Pt,\pi}(x')&\triangleq\sum_{y\in\calX_{h(x')-1}}\sum_{b\in\calA}q^{\Pt,\pi}(y,b)\,\Pt(x'|y,b).
\end{align*}
When $\Pt$ is a transition, $q^{\Pt,\pi}(x,a)$ is the probability of visiting the pair $(x,a)$ in an episode under $\Pt$ and $\pi$, and $q^{\Pt,\pi}(x)$ is that of visiting the state $x$. We also write $q^{\Pt,\pi}(x,a,x')\triangleq q^{\Pt,\pi}(x,a)\Pt(x'|x,a)$ for every $x'\in\calX_{h(x)+1}$, which, when $\Pt$ is a transition, is the probability of visiting the pair $(x,a)$ and then the state $x'$. For a set $\calP$ of transitions, $\Delta(\calP)$ denotes the set of occupancy measures of all transitions in $\calP$ under all policies. As in \citet{jin2020learning}, an element $q\in\Delta(\calP)$ is represented by its values $q(x,a,x')$ on the triples with $x'\in \calX_{h(x)+1}$, with $q(x,a)\triangleq\sum_{x'\in \calX_{h(x)+1}}q(x,a,x')$ and $q(x)\triangleq\sum_{a\in\calA}q(x,a)$, and it determines its policy $\pi^q(a|x)\propto q(x,a)$ and its transition $P^q(x'|x,a)=\frac{q(x,a,x')}{q(x,a)}$ when $q(x,a)>0$, with $P^q(\cdot|x,a)$ an arbitrary distribution over $\calX_{h(x)+1}$ otherwise. All our objectives depend on $q$ only through $q(x,a)$. For a function $f$ on pairs, the value function $V^{\Pt,\pi}(\cdot;f)$ and the action-value function $Q^{\Pt,\pi}(\cdot;f)$ are defined backward from $V^{\Pt,\pi}(x_L;f)\triangleq0$ by
\begin{align*}
Q^{\Pt,\pi}(x,a;f)&\triangleq f(x,a)+\left[\Pt V^{\Pt,\pi}(\cdot;f)\right](x,a),\\
V^{\Pt,\pi}(x;f)&\triangleq\sum_{a\in \calA}\pi(a|x)Q^{\Pt,\pi}(x,a;f),
\end{align*}
so that $\inner{q^{\Pt,\pi},f}=V^{\Pt,\pi}(x_0;f)$. We write $\Pr[\cdot]$ and $\E[\cdot]$ for the probability and the expectation over the randomness of the learner and the environment. With $q_t\triangleq q^{P,\pi_t}$, the regret is
$$\Reg\triangleq\E\left[\sum_{t=1}^T\inner{q_t,\ell_t}\right]-\min_{\pi\in\Pi}\sum_{t=1}^T\inner{q^{P,\pi},\ell_t}.$$
We denote by $\pi^\star$ a minimizer in this definition, that is, the best policy in hindsight, and by $q^\star\triangleq q^{P,\pi^\star}$ its occupancy measure.

\paragraph{Two identities.}
Our derivation uses two elementary identities, which we record here and prove in \pref{app:identities}.

\begin{restatable}{lemma}{rlIdentities}\label{lem:identities}
Let $\Pt$ and $\Ph$ be transitions and $\pi\in\Pi$ a policy. For every function $\varphi$ on states with $\varphi(x_L)=0$,
\begin{align}
\inner{q^{\Pt,\pi},\ \Pt\varphi-\varphi}&=-\varphi(x_0).\label{eq:flow}
\end{align}
Moreover, for every function $f$ on pairs, with $V\triangleq V^{\Pt,\pi}(\cdot;f)$,
\begin{align}
\inner{q^{\Pt,\pi}-q^{\Ph,\pi},\ f}&=\inner{q^{\Ph,\pi},\ (\Pt-\Ph)V}.\label{eq:simulation}
\end{align}
\end{restatable}

The first identity is the \emph{flow conservation}: the function $\Pt\varphi-\varphi$ has the same inner product with every occupancy measure under $\Pt$. The second is the \emph{simulation lemma}: the difference between the losses of a policy under two transitions is a transition error applied to the value function under one of them.

\paragraph{Other notation.}
For distributions $\mu$ and $\nu$ on a finite set $\mathcal{Z}$, $\TV(\mu,\nu)\triangleq\frac12\sum_{z\in\mathcal{Z}}|\mu(z)-\nu(z)|$ is the total variation distance, and $\hsq(\mu,\nu)\triangleq\sum_{z\in\mathcal{Z}}\big(\sqrt{\mu(z)}-\sqrt{\nu(z)}\big)^2$ is the squared Hellinger distance. Direct calculation gives that $\TV(\mu,\nu)\le\sqrt{\hsq(\mu,\nu)}$. For an event $E$, $\indic{E}$ denotes its indicator, which equals $1$ if $E$ occurs and $0$ otherwise.
For a differentiable convex regularizer $\Psi$, we denote by $D_{\Psi}(q,q')\triangleq\Psi(q)-\Psi(q')-\inner{\nabla\Psi(q'),q-q'}$ its Bregman divergence.

\section{Bias-Injected Block Occupancy Mirror Descent}
\label{sec:algorithm}

In this section, we start from UOB-REPS \citep{jin2020learning}, upon which our algorithm is built, and review its regret decomposition to locate where the extra $\sqrt{X}$ factor comes from (\pref{sec:uob-review}). Then, we introduce the general idea of injecting a bias into the vector fed to a blocked occupancy mirror descent (\pref{sec:blocks}), with a detailed design of the bias and the loss estimate in \pref{sec:design-estimators}. Finally, we summarize our algorithm and state its guarantee in \pref{sec:two-track-alg}.

\subsection{Review of UOB-REPS}
\label{sec:uob-review}
We first review UOB-REPS \citep{jin2020learning} and point out its obstacle to achieving $\otil(L\sqrt{XAT})$ regret. UOB-REPS proceeds in epochs indexed by $i$. Denote by $N_i(x,a)$ and $M_i(x'|x,a)$ the numbers of visits to the pair $(x,a)$ and to the triple $(x,a,x')$ before epoch $i$, and by $\bar{P}_i(x'|x,a)\triangleq\frac{M_i(x'|x,a)}{\max\{1,N_i(x,a)\}}$ the empirical transition. The confidence set of epoch $i$ is
\begin{align}
\mathcal{P}_i\triangleq\bigg\{\hat{P}:\ &\left|\hat{P}(x'|x,a)-\bar{P}_i(x'|x,a)\right|\le2\sqrt{\frac{\bar{P}_i(x'|x,a)\iota}{\max\{1,N_i(x,a)-1\}}}+\frac{14\iota}{3\max\{1,N_i(x,a)-1\}}\nonumber\\
&\text{for all }x\in\calX\setminus\{x_L\},\ a\in\calA,\text{ and }x'\in\calX_{h(x)+1}\bigg\},\label{eq:confidence}
\end{align}
where $\hat{P}$ ranges over transitions and $\iota\triangleq14\log(XAT)$. By \citet[Lemma~2]{jin2020learning}, $\Pr\left[P \in \mathcal{P}_i \text{ for all } i\right] \ge 1 - 4(XAT)^{-12}$. In episode $t$ of epoch $i$, UOB-REPS picks $\hat{q}_t \in \Delta(\mathcal{P}_i)$, with policy $\pi_t(a|x) \propto \hat{q}_t(x,a)$ and plausible transition $\hat{P}_t \in \mathcal{P}_i$, that is, the transition in the confidence set such that $\hat{q}_t = q^{\hat{P}_t,\pi_t}$. It computes the upper occupancy bound $u_t(x,a) \triangleq \max_{\hat{P} \in \mathcal{P}_i} q^{\hat{P},\pi_t}(x,a)$ and the optimistic loss estimate $\hat{\ell}_t^{\mathrm{UOB}}(x,a) \triangleq \frac{\ell_t(x,a)\indic{(x,a)\text{ visited in } t}}{u_t(x,a)}$,\footnote{\citet{jin2020learning} add an implicit exploration parameter $\gamma>0$ to the denominator to obtain a high-probability regret bound. Since we only consider the expected regret against an oblivious adversary, we set this parameter to zero in this overview, which simplifies the analysis review below.} and it updates
\begin{align}
\hat{q}_{t+1} &= \mathop{\mathrm{arg\,min}}_{q \in \Delta(\mathcal{P}_i)} \left\{ \langle q, \hat{\ell}_t^{\mathrm{UOB}} \rangle + \frac{1}{\eta} D(q, \hat{q}_t) \right\},\label{eq:uob-update}
\end{align}
where $\eta > 0$ is the learning rate and we use $D$ to denote the relative entropy over state-action pairs.

With the true occupancy measure $q_t$ of $\pi_t$ and the comparator $q^\star$ defined in \pref{sec:prelim}, UOB-REPS decomposes its regret as
\begin{align}
\sum_{t=1}^{T} \langle q_t - q^\star, \ell_t \rangle
&= \underbrace{\sum_{t=1}^{T} \langle q_t - \hat{q}_t, \ell_t \rangle}_{\OccErr}
+ \underbrace{\sum_{t=1}^{T} \langle \hat{q}_t, \ell_t - \hat{\ell}_t^{\mathrm{UOB}} \rangle}_{\LrnBias}
+ \underbrace{\sum_{t=1}^{T} \langle \hat{q}_t - q^\star, \hat{\ell}_t^{\mathrm{UOB}} \rangle}_{\OmdReg}
+ \underbrace{\sum_{t=1}^{T} \langle q^\star, \hat{\ell}_t^{\mathrm{UOB}} - \ell_t \rangle}_{\CompBias}.\label{eq:uob-decomp}
\end{align}
The last two terms are already under control. For nonnegative estimates, the standard analysis of \pref{eq:uob-update} gives
\begin{align}
\sum_{t=1}^{T} \langle \hat{q}_t - q^\star, \hat{\ell}_t^{\mathrm{UOB}} \rangle
&\le \frac{L\log(XA)}{\eta} + \eta \sum_{t=1}^{T} \sum_{x\in \calX\setminus\{x_L\}}\sum_{a\in \calA} \hat{q}_t(x,a)\left(\hat{\ell}_t^{\mathrm{UOB}}(x,a)\right)^2,\label{eq:uob-md}
\end{align}
where the first term is the penalty term and the second is the stability term. Since each term of the stability term has expectation at most $\frac{\hat{q}_t(x,a)}{u_t(x,a)} \le 1$, tuning $\eta$ bounds $\mathbb{E}[\OmdReg]$ by $\widetilde{\order}(\sqrt{LXAT})$, which is even better than $\otil(L\sqrt{XAT})$. Moreover, whenever $P \in \mathcal{P}_i$, we have $u_t \ge q_t$, so $\mathbb{E}_t[\hat{\ell}_t^{\mathrm{UOB}}] \le \ell_t$ and $\mathbb{E}[\CompBias] \le 0$.

The extra factor $\sqrt{X}$ comes from the first two terms, both of which UOB-REPS bounds by the TV distance between the true transition and transitions in the confidence set.
\begin{itemize}
\item For \OccErr{},  \pref{eq:simulation} of \pref{lem:identities} with $(\Pt,\Ph,\pi,f)=(P,\hat P_t,\pi_t,\ell_t)$ gives $\langle q_t - \hat{q}_t, \ell_t \rangle = \langle \hat{q}_t, (P - \hat{P}_t) V_t \rangle$, where $V_t \triangleq V^{P,\pi_t}(\cdot; \ell_t)$ is the value function of $\ell_t$ under $(P,\pi_t)$. Bounding $V_t \le L$ and using
$$\TV\left(P(\cdot|x,a),\hat{P}_t(\cdot|x,a)\right)=\order\left(\sqrt{\frac{|\calX_{h(x)+1}|\iota}{N_i(x,a)}}\right)$$
for every pair $(x,a)$ gives an $\widetilde{\order}(LX\sqrt{AT})$ bound for \OccErr{}.
\item For \LrnBias{}, taking conditional expectations and using $\hat{q}_t \le u_t$ gives
\begin{align*}
\mathbb{E}_t\left[\langle \hat{q}_t, \ell_t - \hat{\ell}_t^{\mathrm{UOB}} \rangle\right]
&= \sum_{x\in \calX\setminus\{x_L\}}\sum_{a\in \calA} \hat{q}_t(x,a)\ell_t(x,a)\frac{u_t(x,a) - q_t(x,a)}{u_t(x,a)} \\
&\le \sum_{x\in \calX\setminus\{x_L\}}\sum_{a\in \calA} \left(u_t(x,a) - q_t(x,a)\right).
\end{align*}
Note that $u_t(x,a)$ is the occupancy measure of $(x,a)$ under a certain transition in the confidence set. Hence, the gap $u_t(x,a) - q_t(x,a)$ is again bounded by the TV distance between that transition and the true one and sums to $\widetilde{\order}(LX\sqrt{AT})$. 
\end{itemize}

\subsection{General Idea: Injecting a Bias over Fixed-Policy Blocks}
\label{sec:blocks}

UOB-REPS bounds the occupancy error by the TV distance between the true transition and transitions in the confidence set, and this term does not involve the loss estimate at all. To reduce it, our idea is to inject a bias into the vector fed to mirror descent. As we will see, this bias has to estimate the value function of the current policy, which is possible only from multiple episodes of that policy. Therefore, we inherit the block update of batch policy evaluation \citep{zhong2023theoretical,viano2024imitation}. We divide the $T$ episodes into $K=\frac{T}{2m}$ blocks of $2m$ consecutive episodes, where block $k\in\{1,\ldots,K\}$ consists of the episodes $\calT_k\triangleq\{2(k-1)m+1,\ldots,2km\}$, hold the policy $\pi_k$ fixed throughout block $k$, and update mirror descent once per block.\footnote{For notational convenience, suppose that $\frac{T}{2m}$ is an integer. Otherwise, the learner plays any policy in the remaining $T-2m\lfloor T/(2m)\rfloor$ episodes, which add at most $2mL\le4X$ to the regret due to the final choice of $m$ shown in \pref{sec:two-track-alg}.} Specifically, from the episodes of block $k$, we construct a loss estimate $\hat{\ell}_k\in\mathbb{R}^{\calX\times \calA}$ together with a \emph{bias} $b_k\in\mathbb{R}^{\calX\times \calA}$, and feed $\hat{d}_k\triangleq\hat{\ell}_k+b_k$ to the mirror descent step. In general, our goal is a refined trade-off among the terms of \pref{eq:uob-decomp}. The bias and the loss estimate bound \OccErr{} and \LrnBias{} by the \emph{squared Hellinger distance} between the true transition and transitions in the confidence set, not by the TV distance, so that their sums over all blocks no longer grow as $\sqrt T$. The price is a mirror descent regret of $\otil(L\sqrt{XAT})$, which is larger than the $\otil(\sqrt{LXAT})$ of UOB-REPS but still matches the lower bound.

To give a general overview of our algorithm, before block $k$, we form the confidence set $\mathcal{P}_k$ by \pref{eq:confidence} with the visit counts and the empirical transition computed from the episodes before $\calT_k$. We then select $\hat{q}_k\in\Delta(\mathcal{P}_k)$ with policy $\pi_k$ and plausible transition $\hat{P}_k\in\mathcal{P}_k$ such that $\hat{q}_k=q^{\hat{P}_k,\pi_k}$, and play the corresponding $\pi_k$ throughout the block. Denote by $\mathcal{F}_{k-1}$ the $\sigma$-field generated before block $k$ and by $\mathbb{E}_k[\cdot]\triangleq\mathbb{E}[\cdot|\mathcal{F}_{k-1}]$, and denote by $\bar{\ell}_k\triangleq\frac{1}{2m}\sum_{t\in\calT_k}\ell_t\in[0,1]^{\calX\times \calA}$ the average loss within block $k$, by $q_k\triangleq q^{P,\pi_k}$ the true occupancy measure of $\pi_k$, and by $V_k\triangleq V^{P,\pi_k}(\cdot;\bar{\ell}_k)\in[0,L]^\calX$ its true value function. With a slight abuse of notation, we use the subscript $k$ for quantities of block $k$ and the subscript $t$ for quantities of episode $t$, as in \pref{sec:uob-review}. After block $k$, we update
\begin{align}
\hat{q}_{k+1}&=\mathop{\mathrm{arg\,min}}_{q\in\Delta(\mathcal{P}_{k+1})}\left\{\langle q,\hat{d}_k\rangle+\frac{1}{\eta}D_{\Psi}(q,\hat{q}_k)\right\},\label{eq:biased-omd}
\end{align}
where $D_{\Psi}$ is the Bregman divergence of the regularizer $\Psi$ to be chosen later.

Since the policy is fixed within each block, the regret can be equivalently written as
\begin{equation*}
\Reg=2m\,\mathbb{E}\left[\sum_{k=1}^K\langle q_k-q^\star,\bar{\ell}_k\rangle\right].
\end{equation*} For each block $k$, we further split $\langle q_k-q^\star,\bar{\ell}_k\rangle=\langle q_k-q^\star,\bar{\ell}_k+\mathbb{E}_k[b_k]\rangle+\langle q^\star-q_k,\mathbb{E}_k[b_k]\rangle$ into the regret on the new loss $\bar{\ell}_k+\mathbb{E}_k[b_k]$ and the difference, and apply \pref{eq:uob-decomp} to the first part with $\hat{d}_k$ as the estimate of the new loss. This gives the following decomposition of the total regret:
\begin{align}
\mathbb{E}\left[\sum_{k=1}^{K}\langle q_k-q^\star,\bar{\ell}_k\rangle\right]
&=\mathbb{E}\Bigg[\underbrace{\sum_{k=1}^{K}\left\langle q_k-\hat{q}_k,\ \bar{\ell}_k+\mathbb{E}_k[b_k]\right\rangle}_{\triangleq\OccErr}
+\underbrace{\sum_{k=1}^{K}\left\langle\hat{q}_k,\ \bar{\ell}_k-\hat{\ell}_k\right\rangle}_{\triangleq\LrnBias}
+\underbrace{\sum_{k=1}^{K}\left\langle\hat{q}_k-q^\star,\ \hat{d}_k\right\rangle}_{\triangleq\OmdReg}\nonumber\\
&\hphantom{{}=\mathbb{E}\Bigg[}+\underbrace{\sum_{k=1}^{K}\left\langle q^\star,\ \hat{\ell}_k-\bar{\ell}_k\right\rangle}_{\triangleq\CompBias}
+\underbrace{\sum_{k=1}^{K}\left\langle q^\star-q_k,\ \mathbb{E}_k[b_k]\right\rangle}_{\triangleq\ShiftPrice}\Bigg],
\label{eq:decomposition}
\end{align}
where we use the fact that $\hat{q}_k$ and $q^\star$ are $\mathcal{F}_{k-1}$-measurable, so that $\mathbb{E}[\langle\hat{q}_k,b_k-\mathbb{E}_k[b_k]\rangle]=\mathbb{E}[\langle q^\star,b_k-\mathbb{E}_k[b_k]\rangle]=0$. The first four terms of \pref{eq:decomposition} are those of \pref{eq:uob-decomp} with respect to the bias-injected loss: the occupancy error \OccErr{}, the learner bias \LrnBias{}, the mirror descent regret \OmdReg{}, and the comparator bias \CompBias{}. The last term \ShiftPrice{} is the negated regret with respect to the bias. In order to improve the guarantee upon UOB-REPS, we aim to design the bias $b_k$ such that \OccErr{} and \ShiftPrice{} are bounded by the squared Hellinger distance, instead of the TV distance as the occupancy error of UOB-REPS. As we will see later, the sum of the squared Hellinger distances will be bounded by terms polylogarithmic in $T$. In addition, we still keep the loss estimate $\hat{\ell}_k$ an optimistic loss under-estimator, similar to that of UOB-REPS, so that \CompBias{} is nonpositive, while making \LrnBias{} bounded by the squared Hellinger distance as well. Finally, as mentioned, we will choose the regularizer $\Psi$ and the block length $m$ to control $2m$ times \OmdReg{} by $\otil(L\sqrt{XAT})$.

\subsection{Designing the Bias and the Loss Estimate}
\label{sec:design-estimators}

We now design the bias $b_k$, which must make \OccErr{} and \ShiftPrice{} bounded by the squared Hellinger distance, and the loss estimate $\hat{\ell}_k$, which must do the same for \LrnBias{} without overestimating the loss.  Throughout this subsection, we assume $P\in\mathcal{P}_k$ for all $k\in[K]$, which holds for all blocks with high probability \citep[Lemma~2]{jin2020learning}.

\paragraph{Bias construction.}
As mentioned, we want to inject a bias $b_k$ such that $\langle q_k - \hat{q}_k, \bar{\ell}_k + \mathbb{E}_k[b_k] \rangle$ is bounded by the squared Hellinger distance, so that its sum over all blocks no longer grows as $\widetilde{\Theta}(LX\sqrt{AT})$. Specifically, let $\widetilde{V}_k^{P'_k} \triangleq V^{P'_k,\pi_k}(\cdot; \bar{\ell}_k)$ be the value function of the loss under a sub-transition $P'_k$ whose next-state distributions are close to those of $P$, such as any transition in $\mathcal{P}_k$. If the bias has expectation $\mathbb{E}_k[b_k] = P\widetilde{V}_k^{P'_k} - \widetilde{V}_k^{P'_k}$, then
\begin{align}
\langle q_k - \hat{q}_k, \bar{\ell}_k + \mathbb{E}_k[b_k] \rangle &= \langle \hat{q}_k, (P - \hat{P}_k)(V_k - \widetilde{V}_k^{P'_k}) \rangle,\label{eq:ideal-bias}
\end{align}
where we use \pref{eq:flow} of \pref{lem:identities} with $\varphi=\widetilde{V}_k^{P'_k}$ under both $P$ and $\hat P_k$, and \pref{eq:simulation} with $(\Pt,\Ph,\pi,f)=(P,\hat P_k,\pi_k,\bar\ell_k)$. Intuitively, $\bar{\ell}_k+P\widetilde{V}_k^{P'_k}-\widetilde{V}_k^{P'_k}$ measures how much worse each action is than the policy $\pi_k$ itself, when the next state follows the true transition and the values afterwards are computed under $P'_k$. That is, the new loss is the advantage of $\pi_k$, obtained by shifting the loss by $P\widetilde{V}_k^{P'_k}-\widetilde{V}_k^{P'_k}$. Note that both $P-\hat{P}_k$ and $V_k-\widetilde{V}_k^{P'_k}$ are controlled by the square root of the confidence widths if $P'_k$ is chosen from $\calP_k$. Hence, by the Cauchy--Schwarz inequality, the right-hand side of \pref{eq:ideal-bias} is bounded by the squared Hellinger distance, and its sum over all blocks is only polylogarithmically dependent on $T$. In addition, this expectation makes \ShiftPrice{} vanish: by  \pref{eq:flow} with $\Pt=P$ and $\varphi=\widetilde{V}_k^{P'_k}$, every occupancy measure under $P$ has the same inner product $-\widetilde{V}_k^{P'_k}(x_0)$ with $P\widetilde{V}_k^{P'_k}-\widetilde{V}_k^{P'_k}$, so $\langle q^\star-q_k,P\widetilde{V}_k^{P'_k}-\widetilde{V}_k^{P'_k}\rangle=0$.

Since $P$ is unknown, the construction of $b_k$ must estimate $P\widetilde{V}_k^{P'_k}$ from the successors observed in block $k$. A natural estimate applies the importance weighting of the UOB-REPS loss estimate to the observed successor values, dividing by the upper occupancy bound $u_k$ instead of the unknown occupancy measure $q_k$:
\begin{align}
\widetilde{b}_k(x,a) &\triangleq \frac{1}{u_k(x,a)}\cdot\frac{1}{m}\sum_{t\in\calT_k} \indic{t\in I_{k,1},\,(x,a)\text{ visited in } t}\,\widetilde{V}_k^{P'_k}(x_{t,h(x)+1}) - \widetilde{V}_k^{P'_k}(x),\label{eq:btilde}
\end{align}
where $u_k(x) \triangleq \max_{\hat{P} \in \mathcal{P}_k} q^{\hat{P},\pi_k}(x)$ is the upper occupancy bound of the state, and $u_k(x,a) \triangleq \pi_k(a|x)u_k(x)$. Here, $I_{k,1}\subseteq\calT_k$ is one of the two halves of block $k$, each of $m$ episodes decided by a fair coin flip. The other half $I_{k,2}$ will be used for the observed losses below. The reason for doing so will be clear later.

However, two concerns remain. First, $\widetilde{V}_k^{P'_k}$ requires the loss $\bar{\ell}_k$, which is unknown. Second, \ShiftPrice{} with this bias is still bounded only by the TV distance between the true transition and transitions in the confidence set. To see this, a direct calculation shows that
\begin{equation*}
\mathbb{E}_k[\widetilde{b}_k](x,a)=[P\widetilde{V}_k^{P'_k}](x,a)-\widetilde{V}_k^{P'_k}(x)-\left(1-\frac{q_k(x)}{u_k(x)}\right)[P\widetilde{V}_k^{P'_k}](x,a).
\end{equation*}
Since $u_k(x)\ge q_k(x)$ whenever $P\in\mathcal{P}_k$, together with $\widetilde{V}_k^{P'_k}\ge0$, we have $\mathbb{E}_k[\widetilde{b}_k](x,a)\le[P\widetilde{V}_k^{P'_k}](x,a)-\widetilde{V}_k^{P'_k}(x)$ for all $(x,a)$. Since $\langle q^\star-q_k,P\widetilde{V}_k^{P'_k}-\widetilde{V}_k^{P'_k}\rangle=0$ by  \pref{eq:flow} of \pref{lem:identities} with $\varphi=\widetilde{V}_k^{P'_k}$, each term in \ShiftPrice{} satisfies
\begin{equation*}
\left\langle q^\star-q_k,\mathbb{E}_k[\widetilde{b}_k]\right\rangle
\le\sum_{x\in \calX\setminus\{x_L\}}\sum_{a\in \calA}q_k(x,a)\left(1-\frac{q_k(x)}{u_k(x)}\right)[P\widetilde{V}_k^{P'_k}](x,a)
\le L\sum_{x\in \calX\setminus\{x_L\}}\left(u_k(x)-q_k(x)\right),
\end{equation*}
which, similar to the learner bias of UOB-REPS, is bounded only by the TV distance.

To address the first concern, we replace the loss $\bar{\ell}_k$ in $\widetilde{V}_k^{P'_k}$ by an estimate computed from the observed losses. Let $O_k(x,a) \triangleq \frac{1}{m}\sum_{t\in\calT_k} \ell_t(x,a)\indic{t\in I_{k,2},\,(x,a)\text{ visited in } t}$ be the average observed loss of $(x,a)$ within the second half $I_{k,2}$ of block $k$. Since the oblivious losses may vary within a block, to make sure that $O_k(x,a)$ is an unbiased estimate of $q_k(x,a)\bar{\ell}_k(x,a)$ for the average loss $\bar{\ell}_k$ of the whole block, as mentioned before, we flip a fair coin to decide whether $I_{k,2}$ is the former or the latter half of $\calT_k$. Let $c_k(x) \triangleq \sum_{a \in \calA} \frac{O_k(x,a)}{u_k(x)+\gamma}$, which is the UOB-REPS estimate of the policy-averaged loss at $x$ with implicit exploration $0<\gamma\le1$. Although we do not aim for a high-probability bound, this implicit exploration plays a different role in controlling the bias scale, which we will explain later. The expectation of $c_k(x)$ is then bounded as follows:
\begin{equation*}
\mathbb{E}_k[c_k(x)]=\frac{q_k(x)}{u_k(x)+\gamma}\sum_{a\in \calA}\pi_k(a|x)\bar{\ell}_k(x,a)\le\sum_{a\in \calA}\pi_k(a|x)\bar{\ell}_k(x,a).
\end{equation*}
Therefore, $c_k(x)$ serves as an optimistic loss estimate of UOB-REPS under $\pi_k$. We then replace $\widetilde{V}_k^{P'_k}$ in $\widetilde{b}_k$ by $V^{P'_k,\pi_k}(\cdot;c_k)$, which uses $c_k$ as the loss.

For the second concern, we choose $P'_k$ to be the lower endpoints of the confidence intervals in \pref{eq:confidence}. Specifically, we define $R_k(x'|x,a)$ as
\begin{align}
R_k(x'|x,a) &\triangleq \max\left\{0,\; \bar{P}_k(x'|x,a) - 2\sqrt{\frac{\bar{P}_k(x'|x,a)\iota}{\max\{1, N_k(x,a)-1\}}} - \frac{14\iota}{3\max\{1, N_k(x,a)-1\}}\right\}.\label{eq:lower-R}
\end{align}
Each $R_k(\cdot|x,a)$ has total mass at most one, so $R_k$ is a sub-transition, and every transition in $\mathcal{P}_k$, including $P$, dominates it entrywise, $R_k\le P$. Hence, $R_k$ is not in the confidence set, and not even a transition, but its next-state distributions are close to those of $P$. We then define the value estimate $\hat{V}_k \triangleq V^{R_k,\pi_k}(\cdot; c_k)$ under $R_k$. Choosing the lower endpoints is the optimism of UOB-REPS applied to the transition: just as the upper occupancy bound $u_k\ge q_k$ makes the loss estimate never overestimate the loss, $R_k\le P$ makes the prediction $[R_k\hat{V}_k](x,a)$ never exceed $[P\hat{V}_k](x,a)$, since $\hat{V}_k\ge0$. Our idea is then to treat $[R_k\hat{V}_k](x,a)$ as a prediction of $[P\hat{V}_k](x,a)$ and to importance-weight only the error of this prediction. That is, in $\widetilde{b}_k$ defined in \pref{eq:btilde}, we replace the successor value $\hat{V}_k(x_{t,h(x)+1})$ by its difference from the prediction, and add the prediction back without importance weighting. This resembles optimistic online learning \citep{rakhlin2013online,wei2018more}. However, there the importance weight is exact and the prediction reduces the variance, while here the importance weight is inexact and the prediction reduces its bias.

Together, our bias is designed as
\begin{align}
b_k(x,a) &\triangleq \frac{S_k(x,a)}{u_k(x,a)} + [R_k\hat{V}_k](x,a) - \hat{V}_k(x),\label{eq:dhat}
\end{align}
where $S_k(x,a) \triangleq \frac{1}{m}\sum_{t\in\calT_k} \indic{t\in I_{k,1},\,(x,a)\text{ visited in } t}\big(\hat{V}_k(x_{t,h(x)+1}) - [R_k\hat{V}_k](x,a)\big)$ is the empirical successor sum. 

Given the independent fair coin and $\mathcal{F}_{k-1}$, the two halves consist of independent trajectories under $(P,\pi_k)$, and the value estimate $\hat{V}_k$ depends only on $I_{k,2}$. Hence, conditioned on $\hat{V}_k$, each episode in $I_{k,1}$ still visits $(x,a)$ with probability $q_k(x,a)$ and then draws $x_{t,h(x)+1}\sim P(\cdot|x,a)$. Denote by $J_k$ the Bernoulli$(1/2)$ random variable representing the outcome of this fair coin. Since $R_k$ is fixed given $\mathcal{F}_{k-1}$, we know that
\begin{align*}
\mathbb{E}_k[S_k(x,a)]&=\mathbb{E}_k\Big[\mathbb{E}_k\big[S_k(x,a)\,\big|\,J_k,\hat{V}_k\big]\Big]=q_k(x,a)\,\mathbb{E}_k\left[\left[(P-R_k)\hat{V}_k\right](x,a)\right]\\
&=q_k(x,a)\left[(P-R_k)\mathbb{E}_k[\hat{V}_k]\right](x,a).
\end{align*}
This is the first reason for splitting block $k$ into two halves: if $\hat{V}_k$ were computed from the same trajectories as $S_k$, the successor state $x_{t,h(x)+1}$ would also enter $\hat{V}_k$, and the conditional expectation above would fail. Since $q_k(x,a)=\pi_k(a|x)q_k(x)$ and $u_k(x,a)=\pi_k(a|x)u_k(x)$, we have $\frac{q_k(x,a)}{u_k(x,a)}=\frac{q_k(x)}{u_k(x)}$, and taking the expectation of \pref{eq:dhat} gives
\begin{align}
\mathbb{E}_k[b_k](x,a) &= \left[P\mathbb{E}_k[\hat{V}_k]\right](x,a) - \mathbb{E}_k[\hat{V}_k](x) - \left(1 - \frac{q_k(x)}{u_k(x)}\right)\left[(P - R_k)\mathbb{E}_k[\hat{V}_k]\right](x,a).\label{eq:td-bias}
\end{align}

This choice makes both \ShiftPrice{} and \OccErr{} bounded by the squared Hellinger distance. Indeed, by  \pref{eq:flow} and  \pref{eq:simulation} in \pref{lem:identities} with $\varphi=\mathbb{E}_k[\hat V_k]$ and $f=\bar\ell_k$, respectively, \pref{eq:td-bias} gives
\begin{align}
\left\langle q^\star-q_k,\mathbb{E}_k[b_k]\right\rangle&\le\sum_{x\in \calX\setminus\{x_L\}}\sum_{a\in \calA}q_k(x,a)\left(1-\frac{q_k(x)}{u_k(x)}\right)\left[(P-R_k)\mathbb{E}_k[\hat{V}_k]\right](x,a),\label{eq:design-shift-price}\\
\left\langle q_k-\hat{q}_k,\bar{\ell}_k+\mathbb{E}_k[b_k]\right\rangle&\le\left\langle\hat{q}_k,(P-\hat{P}_k)\left(V_k-\mathbb{E}_k[\hat{V}_k]\right)\right\rangle\nonumber\\&\hphantom{{}\le{}}+\sum_{x\in \calX\setminus\{x_L\}}\sum_{a\in \calA}\hat{q}_k(x,a)\left(1-\frac{q_k(x)}{u_k(x)}\right)\left[(P-R_k)\mathbb{E}_k[\hat{V}_k]\right](x,a),\label{eq:design-occ-error}
\end{align}
where we drop the terms of the last part of \pref{eq:td-bias} weighted by $q^\star$ and by $q_k$, which are nonpositive by optimism. Each term on the right-hand sides is a product of two gaps. In the two summations of \pref{eq:design-shift-price} and \pref{eq:design-occ-error}, the gaps are $1-\frac{q_k(x)}{u_k(x)}$ and $P-R_k$. In the remaining term, they are $P-\hat{P}_k$ and $V_k-\mathbb{E}_k[\hat{V}_k]$. By linearity of the value function in the loss, $\mathbb{E}_k[\hat{V}_k]=V^{R_k,\pi_k}(\cdot;\mathbb{E}_k[c_k])$, so $V_k-\mathbb{E}_k[\hat{V}_k]$ comes from the gap $P-R_k$ and the gap between $\mathbb{E}_k[c_k]$ and the loss. In \pref{lem:occ-shift}, we show that all these products are bounded by the squared Hellinger distance, up to terms caused by $\gamma$.

\paragraph{Loss estimate construction.}
We now construct the loss estimate $\hat{\ell}_k$, which, together with the bias $b_k$, is fed to the mirror descent update. Our goal is threefold. As in UOB-REPS, it should be an optimistic estimate, $\mathbb{E}_k[\hat{\ell}_k]\le\bar{\ell}_k$, which makes \CompBias{} nonpositive. It should also make \LrnBias{} bounded by the squared Hellinger distance, and it should keep the stability term of \OmdReg{} small.

Since the bias does not appear in \LrnBias{} and \CompBias{}, the first two goals fall on the loss estimate alone. As mentioned in \pref{sec:uob-review}, the learner bias of $\hat{\ell}_t^{\mathrm{UOB}}$ comes from dividing the observed loss by $u_t$ instead of the unknown $q_t$, so it is proportional to the gap $u_t-q_t$, which is bounded only by the TV distance. In block $k$, the same happens if we set the loss estimator as $\frac{O_k(x,a)}{u_k(x,a)}$, whose expectation is $\frac{q_k(x)}{u_k(x)}\bar{\ell}_k(x,a)$. To address this, we multiply it by a correction factor that is linear in an estimate of $q_k(x)$, namely the empirical reach $\tilde{q}_k(x) \triangleq \frac{1}{m}\sum_{t\in\calT_k} \indic{t\in I_{k,1},\,x\text{ visited in } t}$:
\begin{align}
\hat{\ell}_k(x,a)
&\triangleq \frac{O_k(x,a)}{u_k(x,a)}\left(1 + \frac{u_k(x) - \tilde{q}_k(x)}{u_k(x) + \gamma}\right),\label{eq:loss-est}
\end{align}
with the same parameter $\gamma$ as in $c_k$. Note that the correction factor can be negative, since $u_k(x)$ upper bounds $q_k(x)$ but not the empirical reach $\tilde{q}_k(x)$. This is why we also add the implicit exploration parameter $\gamma$ to the loss estimator to control its scale. Since $\tilde{q}_k$ uses $I_{k,1}$ and $O_k$ uses $I_{k,2}$, they are independent given $J_k$ and $\mathcal{F}_{k-1}$, which is the second reason for the split. Together with $\mathbb{E}_k[\tilde{q}_k(x)]=q_k(x)$, this gives, with $\rho_k(x)\triangleq\frac{q_k(x)}{u_k(x)+\gamma}$,
\begin{align}
\mathbb{E}_k[\hat{\ell}_k(x,a)] - \bar{\ell}_k(x,a)
&= \frac{q_k(x)}{u_k(x)}\left(1+\frac{u_k(x)-q_k(x)}{u_k(x)+\gamma}\right)\bar{\ell}_k(x,a)-\bar{\ell}_k(x,a)\nonumber\\
&= -\left(1 - \frac{q_k(x)}{u_k(x)}\right)\left(1 - \rho_k(x)\right)\bar{\ell}_k(x,a) \le 0.\label{eq:loss-bias}
\end{align}
That is, the gap $1-\frac{q_k(x)}{u_k(x)}$ of UOB-REPS is now multiplied by $1-\rho_k(x)$, which is the same gap up to $\gamma$. Thus, each term of \CompBias{} has nonpositive conditional expectation as in UOB-REPS, and, using $\hat{q}_k\le u_k$ and $\bar{\ell}_k\le1$, the learner bias becomes quadratic in the gap up to $\gamma$:
\begin{align}
\mathbb{E}_k\left[\langle\hat{q}_k,\bar{\ell}_k-\hat{\ell}_k\rangle\right]
&\le\sum_{x\in \calX\setminus\{x_L\}}\frac{\left(u_k(x)-q_k(x)\right)^2}{u_k(x)}+\gamma\sum_{x\in \calX\setminus\{x_L\}}\frac{u_k(x)-q_k(x)}{u_k(x)+\gamma}.\label{eq:design-learner-bias}
\end{align}
The first sum is bounded by the squared Hellinger distance, since $\frac{(u_k(x)-q_k(x))^2}{u_k(x)}\le4\big(\sqrt{u_k(x)}-\sqrt{q_k(x)}\big)^2$. In the second sum, $\gamma$ is multiplied by the gap $u_k-q_k$, which will be $o(L\sqrt{XAT})$ after summing over all blocks, as shown in \pref{lem:lrn-comp}.

The third goal is about the whole input $\hat{d}_k = \hat{\ell}_k + b_k$ of mirror descent, and it determines the regularizer $\Psi$, the parameter $\gamma$, and the block length $m$. We start from the analysis of UOB-REPS. Note that the bound \pref{eq:uob-md} requires the estimates to be nonnegative, or at least bounded from below. However, unlike $\hat{\ell}_t^{\mathrm{UOB}}$, $\hat{d}_k$ can be arbitrarily negative without the implicit exploration parameter $\gamma$, as the bias $b_k$ contains $-\hat{V}_k(x)$ and the correction factor in \pref{eq:loss-est} can also be negative, while the denominators $u_k(x)$ in $c_k$ and in \pref{eq:loss-est} can be arbitrarily small. This is where $\gamma$ enters. It bounds the correction factor in \pref{eq:loss-est} and the value estimate. Specifically, we show in \pref{lem:input-size} that for every $(x,a)$, we have
\begin{align*}
\hat q_k(x,a)\left|\hat d_k(x,a)\right|\le\frac{4L}{\gamma}.
\end{align*}

While the scale of the estimator is controlled, regularizing the update using the negative entropy is not enough in order to achieve the target bound, as its stability term, that is, the second term on the right-hand side of \pref{eq:uob-md}, weights the square of each entry of $\hat{d}_k$ only by $\hat{q}_k(x,a)$, which could be too large. Specifically, since $S_k(x,a)$ averages $m$ independent episodes, its second moment is at least $\frac{1}{m}$ times that of a single episode. Hence, with entropy regularizer, the part of the stability term from the successor term $\frac{S_k(x,a)}{u_k(x,a)}$ satisfies
\begin{align}
&\eta\sum_{x\in \calX\setminus\{x_L\}}\sum_{a\in \calA}\hat{q}_k(x,a)\,\mathbb{E}_k\left[\left(\frac{S_k(x,a)}{u_k(x,a)}\right)^2\right]\nonumber\\
&\ge\eta\sum_{x\in \calX\setminus\{x_L\}}\sum_{a\in \calA}\frac{\hat{q}_k(x,a)q_k(x,a)}{m\,u_k(x,a)^2}\sum_{x'\in \calX_{h(x)+1}}P(x'|x,a)\,\mathbb{E}_k\left[\left(\hat{V}_k(x')-[R_k\hat{V}_k](x,a)\right)^2\right].\label{eq:design-entropy-successor}
\end{align}
When $\hat{q}_k(x,a)$ and $q_k(x,a)$ are close to $u_k(x,a)$, each term on the right-hand side is close to $\frac{1}{m}$ times the second moment of the successor value around the prediction, however rarely $(x,a)$ is visited. Since this second moment, $\sum_{x'\in \calX_{h(x)+1}}P(x'|x,a)\,\mathbb{E}_k\big[(\hat{V}_k(x')-[R_k\hat{V}_k](x,a))^2\big]$, can be of order $L^2$ at every pair, the right-hand side can be of order $\frac{\eta XAL^2}{m}$. Summing over the $K$ blocks and multiplying by $2m$, the stability term in $2m$ times \OmdReg{} can then be of order $\frac{\eta TXAL^2}{m}$. Balancing it with the penalty term $\frac{2mL\log(XA)}{\eta}$ gives $\otil(L^{\frac{3}{2}}\sqrt{XAT})$, which will miss the target bound by a factor of $\sqrt{L}$. To address this issue, we add a log-barrier scaled by $\frac{1}{m}$ to the negative entropy \citep{bubeck2018sparsity,lee2020bias}, which gives the following hybrid regularizer:
\begin{align}
\Psi(q) &\triangleq \sum_{x\in \calX\setminus\{x_L\}}\sum_{a\in \calA} \left(q(x,a)\log q(x,a) - \frac{1}{m}\log q(x,a)\right),\label{eq:regularizer}
\end{align}
under which the weight satisfies
\begin{align}
\frac{\hat{q}_k(x,a)^2}{\hat{q}_k(x,a)+1/m}&\le\min\left\{\hat{q}_k(x,a),m\hat{q}_k(x,a)^2\right\}.\label{eq:design-omd-weight}
\end{align}
Now, we are able to bound the stability term in two parts. With the regularizer \pref{eq:regularizer}, the weight $m\hat{q}_k(x,a)^2$ in \pref{eq:design-omd-weight} resolves the issue in \pref{eq:design-entropy-successor}. It cancels the factor $\frac{1}{m}$, and together with $\hat{q}_k\le u_k$, the part of the stability term from the successor term is, up to terms bounded by the confidence widths, at most
\begin{equation*}
\eta\sum_{x\in \calX\setminus\{x_L\}}\sum_{a\in \calA}q_k(x,a)\sum_{x'\in \calX_{h(x)+1}}P(x'|x,a)\,\mathbb{E}_k\left[\left(\hat{V}_k(x')-[R_k\hat{V}_k](x,a)\right)^2\right],
\end{equation*}
whose terms are now proportional to the probability $q_k(x,a)$ of visiting $(x,a)$, however small $u_k(x,a)$ is. With further calculation, we show that the whole stability term, including the parts from the loss estimate and the other terms of the bias, is of order $\eta L(L+\frac{X}{m})$ in each block, up to terms bounded by the confidence widths. See the proof of \pref{lem:omd-regret} in \pref{app:omd} for details. Altogether, the stability term in $2m$ times \OmdReg{} is of order $\eta T(L^2+\frac{LX}{m})$. We then choose the block length to be $m=\lceil\frac{X}{L}\rceil$, which makes $L^2$ and $\frac{LX}{m}$ of the same order, so that the stability term is of order $\eta L^2T$. On the other hand, the log-barrier increases the penalty term from $\otil(\frac{mL}{\eta})$ to $\otil(\frac{XA}{\eta})$. Balancing the two terms by choosing $\eta$  optimally bounds $2m$ times \OmdReg{} by the target $\otil(L\sqrt{XAT})$.

\subsection{Algorithm Overview and Guarantee}
\label{sec:two-track-alg}
We now summarize the full pseudocode of our algorithm, Bias-Injected Block Occupancy Mirror Descent (\BOMD{}), in \pref{alg:main}. It first sets the block length $m=\lceil\frac{X}{L}\rceil$, chosen in \pref{sec:design-estimators}, and the implicit exploration parameter $\gamma=\min\{16mL\eta,1\}$. At the beginning of block $k$, the learner extracts the policy $\pi_k$ from its decision $\hat{q}_k$, as in UOB-REPS, and computes the upper occupancy bounds $u_k$ and the lower endpoints $R_k$ of the current confidence set (\pref{line:extract}). It then plays $\pi_k$ throughout the block and assigns the two halves to $I_{k,1}$ and $I_{k,2}$ using an independent Bernoulli($1/2$) random variable $J_k$ (\pref{line:play}). From $I_{k,1}$, it computes the empirical reaches $\tilde{q}_k$, and from $I_{k,2}$, the observed losses $O_k$ and the optimistic estimate $c_k$, according to \pref{eq:alg-reach}, \pref{eq:alg-obs}, and \pref{eq:alg-c} (\pref{line:estimates}). Next, the algorithm computes the value estimate $\hat{V}_k$ and constructs the successor sums $S_k$ from $I_{k,1}$ according to \pref{eq:alg-succ} (\pref{line:successors}). These form the loss estimate $\hat{\ell}_k$ and the bias $b_k$ (\pref{line:dhat}). Finally, the learner updates the confidence set by \pref{eq:alg-conf} and computes $\hat{q}_{k+1}$ by the mirror descent update \pref{eq:alg-update} with input $\hat{d}_k$ and the hybrid regularizer $\Psi$ defined in \pref{eq:regularizer} (\plines{line:update}{line:omd}). The following theorem gives the guarantee of \pref{alg:main}, which we will prove in \pref{sec:analysis}.

\begin{algorithm}[htbp]
\caption{Bias-Injected Block Occupancy Mirror Descent (\BOMD{})}
\label{alg:main}
\LinesNumbered
\KwIn{Learning rate $\eta>0$.}
Set $m=\lceil \frac{X}{L} \rceil$ and $\gamma=\min\{16mL\eta,1\}$, and initialize $\mathcal{P}_1$ as the set of all transitions and $\hat{q}_1$ as uniform over the state-action pairs of each layer.\;

\For{$k = 1, 2, \dots, K$}{
Extract $\pi_k(a|x) \propto \hat{q}_k(x,a)$, compute the upper occupancy bounds $u_k(x) = \max_{\hat{P} \in \mathcal{P}_k} q^{\hat{P},\pi_k}(x)$ and $u_k(x,a)=\pi_k(a|x)u_k(x)$, and compute the lower endpoints $R_k$ by \pref{eq:lower-R}.\label{line:extract}\;

Play $\pi_k$ in the $2m$ episodes of $\calT_k$, and draw $J_k\sim\mathrm{Bernoulli}(1/2)$ independently. If $J_k=0$, assign the first $m$ episodes to $I_{k,1}$ and the last $m$ to $I_{k,2}$. If $J_k=1$, reverse these assignments.\label{line:play}\;

For every state $x\in \calX\setminus\{x_L\}$ and action $a\in \calA$, compute\label{line:estimates}
\begin{align}
\tilde{q}_k(x)&=\frac{1}{m}\sum_{t\in\calT_k}\indic{t\in I_{k,1},\,x\text{ visited in }t},\label{eq:alg-reach}\\
O_k(x,a)&=\frac{1}{m}\sum_{t\in\calT_k}\ell_t(x,a)\,\indic{t\in I_{k,2},\,(x,a)\text{ visited in }t},\label{eq:alg-obs}\\
c_k(x)&=\sum_{a\in \calA}\frac{O_k(x,a)}{u_k(x)+\gamma}.\label{eq:alg-c}
\end{align}

Compute $\hat{V}_k=V^{R_k,\pi_k}(\cdot\,;c_k)$, and for every state $x\in \calX\setminus\{x_L\}$ and action $a\in \calA$, compute\label{line:successors}
\begin{align}
S_k(x,a)&=\frac{1}{m}\sum_{t\in\calT_k}\indic{t\in I_{k,1},\,(x,a)\text{ visited in }t}\left(\hat{V}_k(x_{t,h(x)+1})-[R_k\hat{V}_k](x,a)\right).\label{eq:alg-succ}
\end{align}\;

Form $\hat{d}_k=\hat{\ell}_k+b_k$ with the loss estimate $\hat{\ell}_k$ in \pref{eq:loss-est} and the bias $b_k$ in \pref{eq:dhat}.\label{line:dhat}

Let $N_{k+1}(x,a)$ and $M_{k+1}(x'|x,a)$ be the numbers of visits to $(x,a)$ and to $(x,a,x')$ before $\calT_{k+1}$, let $\bar{P}_{k+1}(x'|x,a)=\frac{M_{k+1}(x'|x,a)}{\max\{1,N_{k+1}(x,a)\}}$, and form the confidence set\label{line:update}
\begin{align}
\mathcal{P}_{k+1}=\Bigg\{\hat{P}:\ &\left|\hat{P}(x'|x,a)-\bar{P}_{k+1}(x'|x,a)\right|\le2\sqrt{\frac{\bar{P}_{k+1}(x'|x,a)\iota}{\max\{1,N_{k+1}(x,a)-1\}}}\nonumber\\
&\hphantom{\left|\hat{P}(x'|x,a)\right|}+\frac{14\iota}{3\max\{1,N_{k+1}(x,a)-1\}}\text{ for all }(x,a,x')\Bigg\}.\label{eq:alg-conf}
\end{align}

Update\label{line:omd}
\begin{align}
\hat{q}_{k+1}&=\mathop{\mathrm{arg\,min}}_{q \in \Delta(\mathcal{P}_{k+1})} \left\{ \langle q, \hat{d}_k \rangle + \frac{1}{\eta} D_{\Psi}(q, \hat{q}_k) \right\}.\label{eq:alg-update}
\end{align}\;
}

\end{algorithm}

\begin{theorem}
\label{thm:main}
\BOMD{} (\pref{alg:main}) with $\eta=\frac{1}{L}\sqrt{\frac{XA\iota}{T}}$ satisfies
\begin{equation*}
\Reg = \order\left(L\sqrt{XAT\iota} + X^5A\iota^2\right).
\end{equation*}
\end{theorem}

To our knowledge, this is the first efficient algorithm for adversarial MDPs with unknown transitions and bandit feedback whose expected regret matches the $\Omega(L\sqrt{XAT})$ lower bound \citep{jin2018q,domingues2021episodic} up to logarithmic factors and additive lower-order terms. It improves the bounds of UOB-REPS \citep{jin2020learning} and \citet{kulkarni2026introducing} by factors of $\sqrt{X}$ and $L^{\frac{3}{2}}$, respectively.

\section{Analysis}\label{sec:analysis}

In this section, we prove \pref{thm:main}. In \pref{sec:each-term}, we first bound each term of the decomposition \pref{eq:decomposition} on the event that the confidence sets contain the true transition, which holds with high probability, and then in \pref{sec:proof}, we combine these bounds with the contribution of the complement event. The following lemma is an adaptation of Lemma 2 in~\citet{jin2020learning}, whose proof is deferred to \pref{app:analysis}.
\begin{restatable}[{\citealp[Lemma~2]{jin2020learning}}]{lemma}{rlConfidence}\label{lem:confidence}
Let $\calE$ be the event that $P\in\calP_k$ for all $k\in\{1,\ldots,K\}$. Then $\Pr[\calE]\ge1-4(XAT)^{-12}$.
\end{restatable}

Since $\calE$ depends on all blocks, the analysis of block $k$ uses the event $\calG_k\triangleq\{P\in\calP_{k'}\text{ for all }k'\le k\}$, which contains $\calE$ and is determined before block $k$, that is, it is $\calF_{k-1}$-measurable. Hence, the conditional expectations $\E_k[\cdot]$ computed in \pref{sec:algorithm} remain valid on $\calG_k$.

\subsection{Bounding Each Term}\label{sec:each-term}

\paragraph{Confidence widths.}
The analysis of all the terms relies on the size of the confidence sets. For each pair $(x,a)$, let $\epsilon_k(x,a)\triangleq\min\big\{2,\frac{64X\iota}{\max\{1,N_k(x,a)-1\}}\big\}$, which, as shown in \pref{lem:diameter}, upper bounds the squared Hellinger distance between $\Ph(\cdot|x,a)$ and $\Ph'(\cdot|x,a)$ for all $\Ph,\Ph'\in\calP_k$. Let 
\begin{align*}
v_k\triangleq\sum_{x\in\calX\setminus\{x_L\}}\sum_{a\in\calA}\qh_k(x,a)\epsilon_k(x,a)    
\end{align*}
be the width of block $k$ under the decision $\qh_k$, and let $W\triangleq\sum_{k=1}^K\E\big[\indic{\calG_k}v_k\big]$ be the expected cumulative width. The first lemma below shows that the decision $\qh_k$, the upper occupancy bound $u_k$, and the true occupancy measure $q_k$ are close in a square-root sense, at the scale $v_k$ of the squared Hellinger distance. The second shows that $W$ grows only polylogarithmically in $T$. Their proofs are in \pref{app:proof-widths}.

\begin{restatable}{lemma}{rlReach}\label{lem:reach}
\pref{alg:main} guarantees that $\qh_k(x)\le u_k(x)$ for every block $k$ and every state $x\in\calX\setminus\{x_L\}$. If $P\in\calP_k$, then it also guarantees that, for every state $x\in\calX\setminus\{x_L\}$, $q_k(x)\le u_k(x)$, $\big(\sqrt{q_k(x)}-\sqrt{\qh_k(x)}\big)^2\le 8v_k$, and $\big(\sqrt{u_k(x)}-\sqrt{q_k(x)}\big)^2\le 32v_k$. The block width also satisfies $$v_k=\order\left(L\sum_{x\in\calX\setminus\{x_L\}}\sum_{a\in\calA}q_k(x,a)\epsilon_k(x,a)\right).$$
\end{restatable}

\begin{restatable}{lemma}{rlWidths}\label{lem:widths}
\pref{alg:main} guarantees that the expected cumulative width satisfies $W=\order\left(\frac{LX^2A\iota^2}{m}\right)$.
\end{restatable}

In the following, we control each term of \pref{eq:decomposition}.

\paragraph{Bounding \CompBias{} and \LrnBias{}.}
Based on the design of the loss estimate, the following lemma shows that, on the blocks where the confidence sets are correct, \CompBias{} and \LrnBias{} are bounded by $0$ and $\order\left(XW+X\sqrt{K\gamma W}\right)$, respectively.

\begin{restatable}{lemma}{rlLearnerComparator}\label{lem:lrn-comp}
Let $q^\star$ be the comparator defined in \pref{sec:prelim}. If $P\in\calP_k$, then \pref{alg:main} guarantees that
\begin{align*}
\E_k\left[\inner{q^\star,\ellh_k-\ellbar_k}\right]&\le0,\\
\E_k\left[\inner{\qh_k,\ellbar_k-\ellh_k}\right]&\le\order\left(Xv_k+X\sqrt{\gamma v_k}\right).
\end{align*}
Consequently, we have
\begin{align}
\sum_{k=1}^K\E\left[\indic{\calG_k}\inner{q^\star,\ellh_k-\ellbar_k}\right]&\le0,\label{eq:bound-comp}\\
\sum_{k=1}^K\E\left[\indic{\calG_k}\inner{\qh_k,\ellbar_k-\ellh_k}\right]&\le\order\left(XW+X\sqrt{K\gamma W}\right).\label{eq:bound-lrn}
\end{align}
\end{restatable}

Both bounds follow from the expectation \pref{eq:loss-bias} of the loss estimate. The first holds since the loss estimate never overestimates the loss in expectation, and the second since its learner bias is quadratic in the gap between $u_k$ and $q_k$, up to $\gamma$. The proof is in \pref{app:proof-lrn-comp}.

\paragraph{Bounding \OccErr{} and \ShiftPrice{}.}
Next, we control \OccErr{} and \ShiftPrice{}, which depend on the construction of the bias. The following lemma shows that, on the blocks where the confidence sets are correct, both are bounded by $\order(LXW)$, up to a term caused by $\gamma$.

\begin{restatable}{lemma}{rlOccupancyShift}\label{lem:occ-shift}
Let $q^\star$ be the comparator defined in \pref{sec:prelim}. If $P\in\calP_k$, then \pref{alg:main} guarantees that
\begin{align*}
\inner{q^\star-q_k,\E_k[b_k]}&\le\order(LXv_k),\\
\inner{q_k-\qh_k,\ellbar_k+\E_k[b_k]}&\le\order\left(LXv_k+L\sqrt{X\gamma v_k}\right).
\end{align*}
Consequently, we have
\begin{align}
\sum_{k=1}^K\E\left[\indic{\calG_k}\inner{q^\star-q_k,\E_k[b_k]}\right]&\le\order(LXW),\label{eq:bound-shift}\\
\sum_{k=1}^K\E\left[\indic{\calG_k}\inner{q_k-\qh_k,\ellbar_k+\E_k[b_k]}\right]&\le\order\left(LXW+L\sqrt{XK\gamma W}\right).\label{eq:bound-occ}
\end{align}
\end{restatable}

Both bounds start from the expectation \pref{eq:td-bias} of the bias. As discussed in \pref{sec:design-estimators}, each of the two terms is then a sum of products of two gaps, each bounded by the square root of the squared Hellinger distance. The proof is in \pref{app:proof-occ-shift}.

\paragraph{Bounding \OmdReg{}.}
Finally, we control \OmdReg{}, the regret of mirror descent on the inputs $\dhat_k$. Since the log-barrier in $\Psi$ is infinite on the boundary, we compare with a shifted comparator that has positive entries, as in \citet{lee2020bias}. To this end, we define the shifted transition
$$P_0(\cdot|x,a)\triangleq(1-\zeta)P(\cdot|x,a)+\zeta\,\mathrm{Unif}(\calX_{h(x)+1}),$$
where $\zeta\triangleq(XAT)^{-2}$ and $\mathrm{Unif}(\calX_{h(x)+1})$ is the uniform distribution over the next layer, so that every entry of $P_0$ is at least $\frac{\zeta}{X}$. The following lemma, proved in \pref{app:omd}, shows that the confidence sets also contain $P_0$ with high probability.

\begin{restatable}{lemma}{rlShiftedConfidence}\label{lem:confidence-shift}
Let $\calE_0$ be the event that $P_0\in\calP_k$ for all $k\in\{1,\ldots,K\}$. Then $\Pr[\calE_0]\ge1-4(XAT)^{-12}$.
\end{restatable}

Similarly to $\calG_k$, let $\calG_k^0\triangleq\{P_0\in\calP_{k'}\text{ for all }k'\le k\}$, which is $\calF_{k-1}$-measurable. We then define the comparator
$$q^\circ\triangleq\left(1-\frac{1}{T}\right)q^\star+\frac{1}{T}q^{P_0,\pi^{\mathrm{u}}},$$
where $\pi^{\mathrm{u}}$ is the uniform policy. Since every entry of $P_0$ is at least $\frac{\zeta}{X}$, every state is reached with probability at least $\frac{\zeta}{X}$ under $(P_0,\pi^{\mathrm{u}})$, so every entry of $q^\circ$ is at least $\frac{\zeta}{XAT}=(XAT)^{-3}$. Moreover, on $\calG_{k+1}\cap\calG_{k+1}^0$, both $q^\star$ and $q^{P_0,\pi^{\mathrm{u}}}$ lie in $\Delta(\calP_{k+1})$, which is convex since its constraints are linear in $q(x,a,x')$ \citep{jin2020learning}, so $q^\circ\in\Delta(\calP_{k+1})$. The next lemma bounds \OmdReg{} through this comparator.

\begin{restatable}{lemma}{rlOmdRegret}\label{lem:omd-regret}
If $16mL\eta\le1$, then \pref{alg:main} guarantees that
\begin{align}
\E\left[\sum_{k=1}^K\indic{\calG_k}\inner{\qh_k-q^\star,\dhat_k}\right]&\le\frac{LA\iota}{\eta}+\order\left(\eta KL^2+L^2W+\frac{L^2}{m}\right).\label{eq:bound-omd}
\end{align}
\end{restatable}

The complete proof is deferred to \pref{app:omd}. Specifically, we decompose the left-hand side of \pref{eq:bound-omd} into the mirror descent regret against $q^\circ$ on $\calG_k\cap\calG_k^0$, the term involving $q^\circ-q^\star$, and the term on $\calG_k\setminus\calG_k^0$. The last two terms sum to at most $\order(L^2/m)$ by $\|q^\circ-q^\star\|_1\le2L/T$ and \pref{lem:confidence-shift}. The first term requires an involved analysis on the stability term with respect to the hybrid regularizer.

\subsection{Proof of \pref{thm:main}}\label{sec:proof}

We now prove \pref{thm:main} by combining the preceding lemmas. We first consider the case where $16mL\eta>1$. In this case, we have $\gamma=1$ and the choice of $\eta$ in \pref{thm:main} gives $16m\sqrt{XA\iota/T}>1$, or equivalently $T<256m^2XA\iota$. Since each episode has loss at most $L$ and $m\le 2X/L$, we have $\Reg\le LT<256Lm^2XA\iota=\order(X^3A\iota)$, which proves the claimed bound in this case.

We now consider $16mL\eta\le1$. Then $\gamma=16mL\eta$ and \pref{lem:omd-regret} applies. As shown in \pref{sec:blocks}, $\Reg=2m\sum_{k=1}^K\E[\inner{q_k-q^\star,\ellbar_k}]$. Splitting each block according to $\calG_k$, we have
\begin{align*}
\Reg&=2m\sum_{k=1}^K\E\left[\indic{\calG_k}\inner{q_k-q^\star,\ellbar_k}\right]+2m\sum_{k=1}^K\E\left[\indic{\calG_k^c}\inner{q_k-q^\star,\ellbar_k}\right].
\end{align*}
For the second sum, since $\calG_k^c\subseteq\calE^c$, $\inner{q_k-q^\star,\ellbar_k}\le L$, and $2mK=T$, \pref{lem:confidence} gives
\begin{equation*}
2m\sum_{k=1}^K\E\left[\indic{\calG_k^c}\inner{q_k-q^\star,\ellbar_k}\right]\le2mKL\Pr[\calE^c]=LT\Pr[\calE^c]\le4.
\end{equation*}
For the first sum, since $\indic{\calG_k}$, $\qh_k$, and $q^\star$ are $\calF_{k-1}$-measurable, the derivation of \pref{eq:decomposition} applies to each block on $\calG_k$, which gives
\begin{align*}
&\E\left[\indic{\calG_k}\inner{q_k-q^\star,\ellbar_k}\right]\\
&=\E\Big[\indic{\calG_k}\Big(\E_k\left[\inner{q^\star,\ellh_k-\ellbar_k}\right]+\E_k\left[\inner{\qh_k,\ellbar_k-\ellh_k}\right]+\inner{q^\star-q_k,\E_k[b_k]}\\
&\hphantom{{}=\E\Big[\indic{\calG_k}\Big(}+\inner{q_k-\qh_k,\ellbar_k+\E_k[b_k]}+\inner{\qh_k-q^\star,\dhat_k}\Big)\Big].
\end{align*}
Summing over $k$ and multiplying by $2m$, we obtain
\begin{align*}
\Reg&\le4+2m\left(\order\left(XW+X\sqrt{K\gamma W}\right)+\order(LXW)+\order\left(LXW+L\sqrt{XK\gamma W}\right)\right)\\
&\hphantom{{}\le{}}+2m\left(\frac{LA\iota}{\eta}+\order\left(\eta KL^2+L^2W+\frac{L^2}{m}\right)\right)\\
&=4+\frac{2mLA\iota}{\eta}+\order\left(\eta L^2T+L^2+\left(LX+L^2\right)mW+\left(X+L\sqrt{X}\right)\sqrt{mL\eta T\cdot mW}\right)\\
&\le4+\frac{2mLA\iota}{\eta}+\order\left(\eta L^2T+L^2+\left(LX+L^2\right)mW+\frac{m}{L}\left(X^2+L^2X\right)mW\right)\\
&=4+\frac{2mLA\iota}{\eta}+\order\left(\eta L^2T+X^5A\iota^2\right).
\end{align*}
Here, the first inequality applies \pref{eq:bound-comp}, \pref{eq:bound-lrn}, \pref{eq:bound-shift}, \pref{eq:bound-occ}, and \pref{eq:bound-omd} to the five sums in this order. The first equality uses $2mK=T$, $XW\le LXW$, and $2m\sqrt{K\gamma W}=4\sqrt{2mL\eta T\cdot mW}$, which follows from $K=\frac{T}{2m}$ and $\gamma=16mL\eta$. The second inequality uses the AM-GM inequality, which gives
\begin{equation*}
\left(X+L\sqrt{X}\right)\sqrt{mL\eta T\cdot mW}=\sqrt{\eta L^2T\cdot\frac{m}{L}\left(X+L\sqrt{X}\right)^2mW}\le\frac{\eta L^2T}{2}+\frac{m}{2L}\left(X+L\sqrt{X}\right)^2mW
\end{equation*}
and $\left(X+L\sqrt{X}\right)^2\le2X^2+2L^2X$. The last equality uses $mW=\order(LX^2A\iota^2)$ from \pref{lem:widths}, $m\le\frac{2X}{L}$, and $L\le X$. Substituting $\eta=\frac{1}{L}\sqrt{\frac{XA\iota}{T}}$ proves \pref{thm:main}. \hfill$\blacksquare$

\section{Conclusion}\label{sec:conclusion}

In this paper, we study adversarial MDPs with unknown transitions and bandit feedback. We propose \BOMD{}, an efficient algorithm whose expected regret is $\order(L\sqrt{XAT\iota}+X^5A\iota^2)$. To our knowledge, this is the first efficient algorithm that matches the $\Omega(L\sqrt{XAT})$ lower bound up to logarithmic factors and additive lower-order terms. Technically, our algorithm builds on UOB-REPS with several ingredients that may be of independent interest, including a specifically designed bias injected through a value estimate, a block-wise mirror descent update which lets different estimates use separate halves of each block, and a hybrid regularizer.

\appendix
\section{Omitted Details in \pref{sec:prelim}}\label{app:identities}
In this section, we prove the two identities in \pref{lem:identities}.

\restatelemma{lem:identities}{\rlIdentities*}

\begin{proof}
To prove \pref{eq:flow}, denote by $q$ the occupancy measure $q^{\Pt,\pi}$. Since $q$ has unit mass on each layer, and since, for every $x'\in \calX\setminus \calX_0$,
$$\sum_{x\in \calX\setminus\{x_L\}}\sum_{a\in \calA}q(x,a)\Pt(x'|x,a)=q(x'),$$
we have
\begin{align*}
\inner{q,\Pt\varphi}&=\sum_{x\in \calX\setminus \calX_0}q(x)\varphi(x),\\
\inner{q,\varphi}&=\sum_{x\in \calX\setminus\{x_L\}}q(x)\varphi(x).
\end{align*}
Their difference is $q(x_L)\varphi(x_L)-q(x_0)\varphi(x_0)=-\varphi(x_0)$, since $q(x_0)=q(x_L)=1$ and $\varphi(x_L)=0$. This proves \pref{eq:flow}.

To prove \pref{eq:simulation}, observe that the function $f+\Pt V-V=Q^{\Pt,\pi}(\cdot;f)-V$ on pairs averages to zero under $\pi(\cdot|x)$ at every state $x\ne x_L$, so its inner product with $q^{\Ph,\pi}$ vanishes. Hence,
\begin{align*}
\inner{q^{\Ph,\pi},f}&=-\inner{q^{\Ph,\pi},\Ph V-V}-\inner{q^{\Ph,\pi},(\Pt-\Ph)V}\\
&=V(x_0)-\inner{q^{\Ph,\pi},(\Pt-\Ph)V}, &&\text{(by \pref{eq:flow} under $\Ph$ with $\varphi=V$)}
\end{align*}
and $\inner{q^{\Pt,\pi},f}=V(x_0)$ finishes the proof of \pref{eq:simulation}.
\end{proof}

\section{Omitted Details in \pref{sec:analysis}}\label{app:analysis}
In this section, we prove the lemmas used in \pref{sec:analysis}. We first show that the confidence sets contain $P$ at every block with high probability. We then bound the confidence widths and prove the bounds for the terms of \pref{eq:decomposition}.

\restatelemma{lem:confidence}{\rlConfidence*}

\begin{proof}
Apply Lemma 2 of \citet{jin2020learning} with confidence parameter $\delta=(XAT)^{-12}$, which gives $\log(XAT/\delta)=13\log(XAT)<\iota$. The radius in \pref{eq:alg-conf} is therefore larger than the radius in that lemma. The concentration event in their proof holds simultaneously for every state-action pair, successor, and visit count up to $T$. In particular, it holds at the counts $N_k(x,a)$ at the start of every block, giving the claim for the sets in \pref{eq:alg-conf}.
\end{proof}

\subsection{Proofs for Bounding Confidence Widths}\label{app:proof-widths}
In this subsection, we prove \pref{lem:reach} and \pref{lem:widths}. We first use the confidence-set properties in \pref{app:aux} to bound differences in reach probabilities and the width of each block. We then sum the widths over the blocks. We restate the lemmas for convenience.

\restatelemma{lem:reach}{\rlReach*}

\begin{proof}
The definition of $u_k$ gives $\qh_k(x)\le u_k(x)$ since $\Ph_k\in\calP_k$, and likewise $q_k(x)\le u_k(x)$ when $P\in\calP_k$. We assume $P\in\calP_k$ in what follows.

To prove the first inequality, we fix $\Pt,\Ph\in\calP_k$ and denote by $\mu(\tau)$ and $\nu(\tau)$ the probabilities of trajectory $\tau$ from $x_0$ under $(\Pt,\pi_k)$ and $(\Ph,\pi_k)$, respectively. For each $x\in\calX_{h}$, Cauchy--Schwarz gives
\begin{align*}
\sum_{\tau:\,x_{h}=x}\sqrt{\mu(\tau)\nu(\tau)}
&\le\sqrt{\sum_{\tau:\,x_{h}=x}\mu(\tau)}\sqrt{\sum_{\tau:\,x_{h}=x}\nu(\tau)}=\sqrt{q^{\Pt,\pi_k}(x)q^{\Ph,\pi_k}(x)}.
\end{align*}
Expanding the squares and summing over $x\in\calX_{h}$, we obtain
\begin{align}
\sum_{x\in\calX_{h}}\left(\sqrt{q^{\Pt,\pi_k}(x)}-\sqrt{q^{\Ph,\pi_k}(x)}\right)^2&\le\sum_\tau\left(\sqrt{\mu(\tau)}-\sqrt{\nu(\tau)}\right)^2\nonumber\\
&\le\frac{15}{2}\sum_{y\in\calX\setminus\{x_L\}}\sum_{a\in\calA}q^{\Pt,\pi_k}(y,a)\hsq\left(\Pt(\cdot|y,a),\Ph(\cdot|y,a)\right)\nonumber\\
&\le\frac{15}{2}\sum_{y\in\calX\setminus\{x_L\}}\sum_{a\in\calA}q^{\Pt,\pi_k}(y,a)\epsilon_k(y,a),\label{eq:app-reach-comparison}
\end{align}
where the second inequality uses \pref{lem:path} with starting state $x_0$, and the third uses \pref{lem:diameter}(a) for each $(y,a)$. Taking $(\Pt,\Ph)=(\Ph_k,P)$ in \pref{eq:app-reach-comparison} for the layer containing $x$ bounds $\left(\sqrt{\qh_k(x)}-\sqrt{q_k(x)}\right)^2$ by $8v_k$.

To prove the second inequality, we choose $P^x\in\calP_k$ such that $q^{P^x,\pi_k}(x)=u_k(x)$. Applying \pref{eq:app-reach-comparison} with $(\Pt,\Ph)=(\Ph_k,P^x)$ gives $\left|\sqrt{u_k(x)}-\sqrt{\qh_k(x)}\right|\le\sqrt{8v_k}$. Combining the two applications of \pref{eq:app-reach-comparison} by the triangle inequality gives $\left|\sqrt{u_k(x)}-\sqrt{q_k(x)}\right|\le2\sqrt{8v_k}$.

Finally, to bound $v_k$, we use $\qh_k(y)\le2q_k(y)+2\left(\sqrt{\qh_k(y)}-\sqrt{q_k(y)}\right)^2$ and $\sum_{a\in\calA}\pi_k(a|y)\epsilon_k(y,a)\le2$. Then
\begin{align}
v_k&\le 2\sum_{y\in\calX\setminus\{x_L\}}\sum_{a\in\calA}q_k(y,a)\epsilon_k(y,a)
+4\sum_{h=0}^{L-1}\sum_{y\in\calX_{h}}\left(\sqrt{\qh_k(y)}-\sqrt{q_k(y)}\right)^2\nonumber\\
&\le(2+30L)\sum_{y\in\calX\setminus\{x_L\}}\sum_{a\in\calA}q_k(y,a)\epsilon_k(y,a),\label{eq:app-width-comparison}
\end{align}
where the second inequality applies \pref{eq:app-reach-comparison} with $(\Pt,\Ph)=(P,\Ph_k)$ to each of the $L$ layers, giving $4\cdot(15/2)L=30L$.
\end{proof}

\restatelemma{lem:widths}{\rlWidths*}

\begin{proof}
By the definition of $W$ and \pref{eq:app-width-comparison} in the proof of \pref{lem:reach}, we have
\begin{align*}
W&=\sum_{k=1}^K\E\left[\indic{\calG_k}v_k\right]\\
&\le(2+30L)\sum_{k=1}^K\E\left[\indic{\calG_k}\sum_{x\in\calX\setminus\{x_L\}}\sum_{a\in\calA}q_k(x,a)\epsilon_k(x,a)\right]\\
&=\frac{2+30L}{2m}\sum_{k=1}^K\E\left[\indic{\calG_k}\sum_{x\in\calX\setminus\{x_L\}}\sum_{a\in\calA}\left(N_{k+1}(x,a)-N_k(x,a)\right)\epsilon_k(x,a)\right].
\end{align*}
The last equality uses $\E_k[N_{k+1}(x,a)-N_k(x,a)]=2m q_k(x,a)$, since the learner plays $\pi_k$ for all $2m$ episodes and both $\indic{\calG_k}$ and $\epsilon_k(x,a)$ are determined before block $k$.

For each $(x,a)$, the blocks with $N_k(x,a)\le2m$ contain at most $4m$ visits in total, since a block adds at most $2m$ visits. Their contribution is at most $8m$ by $\epsilon_k(x,a)\le2$. When $N_k(x,a)>2m$, we have $N_{k+1}(x,a)\le2N_k(x,a)$ and $\epsilon_k(x,a)\le 128X\iota/N_k(x,a)$. Hence,
\begin{align*}
\sum_{k:\,N_k(x,a)>2m}\frac{N_{k+1}(x,a)-N_k(x,a)}{N_k(x,a)}
&\le2\sum_{k:\,N_k(x,a)>2m}\log\frac{N_{k+1}(x,a)}{N_k(x,a)}\le2\log T,
\end{align*}
where we used $y-1\le2\log y$ for $1\le y\le2$ and telescoped the logarithms. Combining the two cases gives
\begin{align*}
\sum_{k=1}^K\left(N_{k+1}(x,a)-N_k(x,a)\right)\epsilon_k(x,a)&\le 8m+\order(X\iota\log T)=\order(X\iota^2).
\end{align*}
Since $\indic{\calG_k}\le1$, summing over $(x,a)$ in the bound on $W$ gives $W=\order(LX^2A\iota^2/m)$.
\end{proof}

\subsection{Proofs for Bounding \CompBias{} and \LrnBias{}}\label{app:proof-lrn-comp}
In this subsection, we prove \pref{lem:lrn-comp}, which will be useful in bounding \CompBias{} and \LrnBias{}.

\restatelemma{lem:lrn-comp}{\rlLearnerComparator*}

\begin{proof}
For the first inequality, \pref{eq:loss-bias} gives $\E_k[\ellh_k(x,a)]\le\ellbar_k(x,a)$ for every $(x,a)$. Taking the inner product with $q^\star\ge0$ gives $\E_k[\inner{q^\star,\ellh_k-\ellbar_k}]\le0$.

For the second inequality, we use \pref{lem:reach} to bound $\sqrt{u_k(x)}-\sqrt{q_k(x)}$ by $\sqrt{32v_k}$ and use $q_k(x)\le u_k(x)$ to bound $\sqrt{u_k(x)}+\sqrt{q_k(x)}$ by $2\sqrt{u_k(x)}$. Thus,
\begin{align*}
u_k(x)-q_k(x)&=\left(\sqrt{u_k(x)}-\sqrt{q_k(x)}\right)\left(\sqrt{u_k(x)}+\sqrt{q_k(x)}\right)\le2\sqrt{32u_k(x)v_k}.
\end{align*}
We square this inequality to bound the first term of \pref{eq:design-learner-bias} and use $u_k(x)+\gamma\ge2\sqrt{u_k(x)\gamma}$ in the second. Specifically, we have
\begin{align}
\frac{\left(u_k(x)-q_k(x)\right)^2}{u_k(x)}&\le128v_k,\label{eq:app-learner-gap}\\
\gamma\frac{u_k(x)-q_k(x)}{u_k(x)+\gamma}&\le\frac{2\gamma\sqrt{32u_k(x)v_k}}{u_k(x)+\gamma}\le\sqrt{32\gamma v_k}.\nonumber
\end{align}
Substituting into \pref{eq:design-learner-bias} gives $\E_k[\inner{\qh_k,\ellbar_k-\ellh_k}]\le128Xv_k+X\sqrt{32\gamma v_k}$. We multiply the two block inequalities by $\indic{\calG_k}$ and take expectations. Summing over $k$ gives \pref{eq:bound-comp} and \pref{eq:bound-lrn}, since Cauchy--Schwarz gives $\sum_{k=1}^K\E[\indic{\calG_k}\sqrt{v_k}]\le\sqrt{KW}$.
\end{proof}

\subsection{Proofs for Bounding \OccErr{} and \ShiftPrice{}}\label{app:proof-occ-shift}
In this subsection, we prove \pref{lem:occ-shift} by bounding the remainder in \pref{eq:td-bias} and the product of transition and value estimation errors in the occupancy error.

\restatelemma{lem:occ-shift}{\rlOccupancyShift*}

\begin{proof}
We first bound the shift price using \pref{eq:design-shift-price}. By \pref{lem:diameter}(c), $(P-R_k)(x'|x,a)\ge0$ for every $x'$, and its sum over $x'$ is $\delta_k(x,a)\triangleq1-\sum_{x'\in\calX_{h(x)+1}}R_k(x'|x,a)\le\sqrt{\epsilon_k(x,a)}$. Since $0\le\E_k[\Vh_k(x')]\le L$, we have $[(P-R_k)\E_k[\Vh_k]](x,a)\le L\delta_k(x,a)$. By \pref{eq:design-shift-price} and Cauchy--Schwarz, we obtain
\begin{align*}
\inner{q^\star-q_k,\E_k[b_k]}&\le L\sum_{x\in\calX\setminus\{x_L\}}\sum_{a\in\calA}q_k(x,a)\left(1-\frac{q_k(x)}{u_k(x)}\right)\delta_k(x,a)\\
&\le L\sqrt{\sum_{x\in\calX\setminus\{x_L\}} q_k(x)\left(1-\frac{q_k(x)}{u_k(x)}\right)^2}\sqrt{\sum_{x\in\calX\setminus\{x_L\}}\sum_{a\in\calA}q_k(x,a)\epsilon_k(x,a)}.
\end{align*}
For the first factor, \pref{eq:app-learner-gap} gives $u_k(x)(1-q_k(x)/u_k(x))^2\le128v_k$, and $q_k(x)\le u_k(x)$. For the second, \pref{lem:reach} gives $q_k(x)\le2\qh_k(x)+16v_k$. Since $\sum_{a\in\calA}\pi_k(a|x)\epsilon_k(x,a)\le2$, we obtain
\begin{align*}
\sum_{x\in\calX\setminus\{x_L\}}\sum_{a\in\calA}q_k(x,a)\epsilon_k(x,a)&\le2v_k+16v_k\sum_{x\in\calX\setminus\{x_L\}}\sum_{a\in\calA}\pi_k(a|x)\epsilon_k(x,a)\le(2+32X)v_k.
\end{align*}
Thus $\inner{q^\star-q_k,\E_k[b_k]}\le L\sqrt{128Xv_k}\sqrt{(2+32X)v_k}=\order(LXv_k)$.

We next bound the occupancy error using \pref{eq:design-occ-error}. By Cauchy--Schwarz, \pref{eq:app-learner-gap}, $\qh_k(x)\le u_k(x)$, \pref{lem:diameter}(c), and the definition of $v_k$, we have
\begin{align}
&\sum_{x\in\calX\setminus\{x_L\}}\sum_{a\in\calA}\qh_k(x,a)\left(1-\frac{q_k(x)}{u_k(x)}\right)[(P-R_k)\E_k[\Vh_k]](x,a)\nonumber\\
&\le L\sqrt{\sum_{x\in\calX\setminus\{x_L\}}\qh_k(x)\left(1-\frac{q_k(x)}{u_k(x)}\right)^2}\sqrt{\sum_{x\in\calX\setminus\{x_L\}}\sum_{a\in\calA}\qh_k(x,a)\epsilon_k(x,a)}\le L\sqrt{128X}\,v_k.\label{eq:app-occ-last-term}
\end{align}
Substituting into \pref{eq:design-occ-error}, we obtain
\begin{align*}
\inner{q_k-\qh_k,\ellbar_k+\E_k[b_k]}&\le\inner{\qh_k,(P-\Ph_k)(V_k-\E_k[\Vh_k])}+L\sqrt{128X}\,v_k.
\end{align*}
On $\calG_k$, every function $\varphi$ on states with $\varphi(x_L)=0$ and $\left|\varphi\right|\le L$ satisfies
\begin{align}
\left|\inner{\qh_k,(P-\Ph_k)\varphi}\right|&=\left|\sum_{h=0}^{L-1}\sum_{x\in\calX_h}\sum_{a\in\calA}\sum_{z\in\calX_{h+1}}\qh_k(x,a)\left(P(z|x,a)-\Ph_k(z|x,a)\right)\varphi(z)\right|\nonumber\\
&\le Lv_k+2\left(\sum_{x\in\calX\setminus\{x_L\}}\sum_{a\in\calA}\qh_k(x,a)\hsq\left(P(\cdot|x,a),\Ph_k(\cdot|x,a)\right)\right)^{1/2}\nonumber\\
&\hphantom{{}\le Lv_k+2{}}\cdot\left(\sum_{h=0}^{L-1}\sum_{z\in\calX_{h+1}}\varphi(z)^2\sum_{x\in\calX_h}\sum_{a\in\calA}\qh_k(x,a)\Ph_k(z|x,a)\right)^{1/2}\nonumber\\
&\le Lv_k+2\sqrt{v_k\sum_{z\in\calX\setminus\{x_0\}}\qh_k(z)\varphi(z)^2}\nonumber\\
&\le Lv_k+2\sqrt{v_k\sum_{z\in\calX\setminus\{x_L\}}\qh_k(z)\varphi(z)^2},\label{eq:app-transition-test}
\end{align}
where the first inequality uses $P-\Ph_k=(\sqrt P-\sqrt{\Ph_k})^2+2\sqrt{\Ph_k}(\sqrt P-\sqrt{\Ph_k})$ entrywise and Cauchy--Schwarz, the first two inequalities use \pref{lem:diameter}(a) and the definition of $v_k$, and the last uses $\varphi(x_L)=0$. We apply \pref{eq:app-transition-test} to $\varphi=V_k-\E_k[\Vh_k]$.

To further bound the second term on the right-hand side of \pref{eq:app-transition-test}, we bound $\sum_{z\in\calX\setminus\{x_L\}}\qh_k(z)(V_k(z)-\E_k[\Vh_k(z)])^2$. Since each episode belongs to $I_{k,2}$ with probability $1/2$ independently of its trajectory, $\E_k[O_k(x,a)]=q_k(x,a)\ellbar_k(x,a)$. Substituting this into the definition of $c_k$ gives
\begin{align*}
\E_k[c_k(x)]=\frac{\sum_{a\in\calA}\E_k[O_k(x,a)]}{u_k(x)+\gamma}
=\frac{\sum_{a\in\calA}q_k(x,a)\ellbar_k(x,a)}{u_k(x)+\gamma}=\rho_k(x)\sum_{a\in\calA}\pi_k(a|x)\ellbar_k(x,a),
\end{align*}
where the last equality uses $q_k(x,a)=q_k(x)\pi_k(a|x)$ and $\rho_k(x)=q_k(x)/(u_k(x)+\gamma)$. Since $R_k$ and $\pi_k$ are fixed before block $k$, the Bellman recursion for $\Vh_k$ gives
\begin{align*}
\E_k[\Vh_k(x)]&=\E_k[c_k(x)]+\sum_{a\in\calA}\pi_k(a|x)[R_k\E_k[\Vh_k]](x,a)\\
&=\sum_{a\in\calA}\pi_k(a|x)\left(\rho_k(x)\ellbar_k(x,a)+[R_k\E_k[\Vh_k]](x,a)\right).
\end{align*}
Subtracting this expression from the Bellman recursion for $V_k$ yields
\begin{align*}
V_k(x)-\E_k[\Vh_k(x)]
&=\sum_{a\in\calA}\pi_k(a|x)\left((1-\rho_k(x))\ellbar_k(x,a)+[PV_k](x,a)-[R_k\E_k[\Vh_k]](x,a)\right)\\
&=\sum_{a\in\calA}\pi_k(a|x)\left((1-\rho_k(x))\ellbar_k(x,a)+[(P-R_k)V_k](x,a)\right)\\
&\hphantom{{}={}}+\sum_{a\in\calA}\pi_k(a|x)[R_k(V_k-\E_k[\Vh_k])](x,a),
\end{align*}
where the second equality adds and subtracts $[R_kV_k](x,a)$.
By \pref{lem:diameter}(c) and $0\le V_k\le L$, the quantity inside the first pair of parentheses satisfies
\begin{align*}
0\le(1-\rho_k(x))\ellbar_k(x,a)+[(P-R_k)V_k](x,a)&\le1-\rho_k(x)+L\delta_k(x,a).
\end{align*}
Both values vanish at $x_L$. Unrolling the recursion from $z$ and using $0\le R_k\le\Ph_k$ gives
\begin{align*}
0\le V_k(z)-\E_k[\Vh_k(z)]
&=V^{R_k,\pi_k}\left(z;(1-\rho_k)\ellbar_k+(P-R_k)V_k\right)\\
&\le V^{R_k,\pi_k}\left(z;1-\rho_k+L\delta_k\right)\le V^{\Ph_k,\pi_k}\left(z;1-\rho_k+L\delta_k\right).
\end{align*}
For a function $f$ on state-action pairs, denote by $f^2$ its coordinatewise square, so $f^2(x,a)\triangleq f(x,a)^2$. Starting from $z$, the sum of the conditional occupancies is $L-h(z)$. Cauchy--Schwarz with these weights gives
\begin{align*}
V^{\Ph_k,\pi_k}(z;f)^2
&=\left(\sum_{h'=h(z)}^{L-1}\sum_{x\in\calX_{h'}}\sum_{a\in\calA}q^{\Ph_k,\pi_k}(x,a|z)f(x,a)\right)^2\\
&\le (L-h(z))\sum_{h'=h(z)}^{L-1}\sum_{x\in\calX_{h'}}\sum_{a\in\calA}q^{\Ph_k,\pi_k}(x,a|z)f(x,a)^2\le L V^{\Ph_k,\pi_k}(z;f^2).
\end{align*}
For $x\in\calX_{h'}$ and $h\le h'$, we have $\sum_{z\in\calX_{h}}\qh_k(z)q^{\Ph_k,\pi_k}(x,a|z)=\qh_k(x,a)$ by conditioning on the state at layer $h$. Summing over the starting layers gives
\begin{align}
\sum_{z\in\calX\setminus\{x_L\}}\qh_k(z)V^{\Ph_k,\pi_k}(z;f)^2
&\le L\sum_{h=0}^{L-1}\sum_{h'=h}^{L-1}\sum_{x\in\calX_{h'}}\sum_{a\in\calA}\qh_k(x,a)f(x,a)^2\nonumber\\
&\le L^2\sum_{x\in\calX\setminus\{x_L\}}\sum_{a\in\calA}\qh_k(x,a)f(x,a)^2.\label{eq:app-value-second-moment}
\end{align}
We apply \pref{eq:app-value-second-moment} with $f(x,a)=1-\rho_k(x)+L\delta_k(x,a)$. Since $1-\rho_k(x)=(u_k(x)-q_k(x)+\gamma)/(u_k(x)+\gamma)$, we have
\begin{align*}
\qh_k(x)(1-\rho_k(x))^2
&\le 2\frac{(u_k(x)-q_k(x))^2}{u_k(x)}+2\frac{u_k(x)\gamma^2}{(u_k(x)+\gamma)^2}\le 256v_k+\gamma/2,
\end{align*}
where the last inequality uses \pref{eq:app-learner-gap} and $(u_k(x)+\gamma)^2\ge4u_k(x)\gamma$. Thus $(\alpha+\beta)^2\le2\alpha^2+2\beta^2$ gives
\begin{align}
&\sum_{z\in\calX\setminus\{x_L\}}\qh_k(z)\left(V_k(z)-\E_k[\Vh_k(z)]\right)^2\nonumber\\
&\le 2L^2\left(\sum_{x\in\calX\setminus\{x_L\}}\qh_k(x)(1-\rho_k(x))^2+L^2\sum_{x\in\calX\setminus\{x_L\}}\sum_{a\in\calA}\qh_k(x,a)\delta_k(x,a)^2\right)\nonumber\\
&\le 2L^2\left(X(256v_k+\gamma/2)+L^2v_k\right),\label{eq:app-value-error}
\end{align}
where the last inequality uses $\delta_k(x,a)^2\le\epsilon_k(x,a)$ and the definition of $v_k$. Substituting \pref{eq:app-value-error} into \pref{eq:app-transition-test} gives
\begin{align*}
\left|\inner{\qh_k,(P-\Ph_k)(V_k-\E_k[\Vh_k])}\right|
&\le Lv_k+2L\sqrt{2v_k\left(X(256v_k+\gamma/2)+L^2v_k\right)}\\
&\le\order\left(L(\sqrt{X}+L)v_k+L\sqrt{X\gamma v_k}\right).
\end{align*}
Substituting this inequality and \pref{eq:app-occ-last-term} into \pref{eq:design-occ-error} gives
\begin{align*}
\inner{q_k-\qh_k,\ellbar_k+\E_k[b_k]}&\le\order\left(LXv_k+L\sqrt{X\gamma v_k}\right),
\end{align*}
where we used $\sqrt X\le X$ and $L\le X$. Finally, we multiply the shift-price and occupancy-error bounds by $\indic{\calG_k}$, take expectations, and sum over $k$. Using $\sum_{k=1}^K\E[\indic{\calG_k}\sqrt{v_k}]\le\sqrt{KW}$ gives \pref{eq:bound-shift} and \pref{eq:bound-occ}, respectively.
\end{proof}

\subsection{Proofs for Bounding \OmdReg{}}\label{app:omd}
In this subsection, we prove \pref{lem:confidence-shift} and \pref{lem:omd-regret}. The first lemma makes the shifted comparator feasible on the confidence event. For the second, we bound the mirror descent input and apply a one-step inequality for the hybrid regularizer.

\restatelemma{lem:confidence-shift}{\rlShiftedConfidence*}

\begin{proof}
On the concentration event used in the proof of \pref{lem:confidence}, which has probability at least $1-4(XAT)^{-12}$, $P$ indeed satisfies \pref{eq:alg-conf} in every block with $13\log(XAT)$ in place of $\iota=14\log(XAT)$. By the definition of $P_0$ and $N_k(x,a)\le T$, for every $(x,a,x')$ we have
\begin{align}
\left|P_0(x'|x,a)-P(x'|x,a)\right|&\le\zeta=(XAT)^{-2}\le\frac{14\log(XAT)}{3\max\{1,N_k(x,a)-1\}}.\label{eq:app-shift-size}
\end{align}
On this event, the triangle inequality gives
\begin{align*}
\left|P_0(x'|x,a)-\Pb_k(x'|x,a)\right|
&\le \left|P(x'|x,a)-\Pb_k(x'|x,a)\right|+\left|P_0(x'|x,a)-P(x'|x,a)\right|\\
&\le 2\sqrt{\frac{13\log(XAT)\Pb_k(x'|x,a)}{\max\{1,N_k(x,a)-1\}}}+\frac{14\left(13\log(XAT)+\log(XAT)\right)}{3\max\{1,N_k(x,a)-1\}}\\
&\le 2\sqrt{\frac{\Pb_k(x'|x,a)\iota}{\max\{1,N_k(x,a)-1\}}}+\frac{14\iota}{3\max\{1,N_k(x,a)-1\}},
\end{align*}
where the second inequality uses the bound for $P$ and \pref{eq:app-shift-size}, and the last uses $\iota=14\log(XAT)$. Thus, by \pref{eq:alg-conf}, $P_0\in\calP_k$ for every block $k$ on the same event.
\end{proof}

The following lemma bounds the scale of $\wh{d}_k$, which will be useful in the proof of \pref{lem:omd-regret}.

\begin{lemma}\label{lem:input-size}
If $\gamma\le1$, then for every block $k$, every state $x\in\calX\setminus\{x_L\}$, and every action $a\in\calA$, \pref{alg:main} guarantees that
\begin{align*}
\qh_k(x,a)\left|\dhat_k(x,a)\right|\le\frac{4L}{\gamma}.
\end{align*}
\end{lemma}

\begin{proof}
Since $\qh_k\in\Delta(\calP_k)$ and its policy is $\pi_k$, the definition of $u_k$ gives $\qh_k(x)\le u_k(x)$, and hence $\hat q_k(x,a)=\pi_k(a|x)\qh_k(x)\le u_k(x,a)$. We first bound the two terms of $b_k$ in \pref{eq:dhat}. Since $\sum_{a\in\calA}O_k(x,a)\le1$, we have $0\le c_k(x)\le1/\gamma$. The value under $R_k$ sums at most $L$ such losses, and $R_k(\cdot|x,a)$ has mass at most one. Thus both $\hat V_k(x')$ and $[R_k\hat V_k](x,a)$ lie in $[0,L/\gamma]$. By the definition of $S_k$, $\hat q_k(x,a)\le u_k(x,a)$, and $\hat q_k(x,a)\le1$, we obtain
\begin{align*}
\hat q_k(x,a)\left|\frac{S_k(x,a)}{u_k(x,a)}\right|
&\le\frac{1}{m}\sum_{t\in I_{k,1}}\indic{(x,a)\text{ visited in }t}\left|\hat V_k(x_{t,h(x)+1})-[R_k\hat V_k](x,a)\right|\le\frac{L}{\gamma},\\
\hat q_k(x,a)\left|[R_k\hat V_k](x,a)-\hat V_k(x)\right|&\le\hat q_k(x,a)\frac{L}{\gamma}\le\frac{L}{\gamma}.
\end{align*}
For the loss estimate in \pref{eq:loss-est}, $O_k(x,a)\le1$ and $u_k(x),\tilde q_k(x)\in[0,1]$. Since $\gamma\le1$, we have
\begin{align*}
\hat q_k(x,a)|\hat\ell_k(x,a)|
&\le\frac{\hat q_k(x,a)}{u_k(x,a)}O_k(x,a)\left(1+\frac{|u_k(x)-\tilde q_k(x)|}{u_k(x)+\gamma}\right)\le 1+\frac1\gamma\le\frac{2}{\gamma}.
\end{align*}
Adding the three bounds and using $L\ge1$ gives, for every $(x,a)$,
\begin{align*}
\hat q_k(x,a)\left|\hat d_k(x,a)\right|&\le\frac{2}{\gamma}+\frac{L}{\gamma}+\frac{L}{\gamma}\le\frac{4L}{\gamma},
\end{align*}
finishing the proof.
\end{proof}

The next lemma bounds the regret of the mirror descent updates on the bias-injected losses.

\restatelemma{lem:omd-regret}{\rlOmdRegret*}

\begin{proof}
We first decompose the left-hand side of \pref{eq:bound-omd}. As defined in \pref{sec:analysis} and \pref{sec:each-term}, $\calG_k$ and $\calG_k^0$ are the events that every confidence set through block $k$ contains $P$ and $P_0$, respectively. Both are $\calF_{k-1}$-measurable. Adding and subtracting $q^\circ$ gives
\begin{align}
&\E\left[\sum_{k=1}^K\indic{\calG_k}\inner{\qh_k-q^\star,\dhat_k}\right]\nonumber\\
&=\sum_{k=1}^K\E\left[\indic{\calG_k\cap\calG_k^0}\inner{\qh_k-q^\circ,\dhat_k}\right]\nonumber\\
&\hphantom{{}={}}+\sum_{k=1}^K\E\left[\indic{\calG_k}\inner{q^\circ-q^\star,\E_k[\dhat_k]}\right]+\sum_{k=1}^K\E\left[\indic{\calG_k\setminus\calG_k^0}\inner{\qh_k-q^\circ,\E_k[\dhat_k]}\right].\label{eq:app-comparator-decomposition}
\end{align}
To bound the last two terms, we use \pref{eq:loss-bias}, which gives $0\le\E_k[\ellh_k]\le\ellbar_k\le1$ on $\calG_k$. By \pref{eq:td-bias},
\begin{align*}
\left|\E_k[b_k(x,a)]\right|
&\le\left|[P\E_k[\Vh_k]](x,a)-\E_k[\Vh_k(x)]\right|+\left(1-\frac{q_k(x)}{u_k(x)}\right)[(P-R_k)\E_k[\Vh_k]](x,a)\le2L,\\
\left\|\E_k[\dhat_k]\right\|_\infty&\le\left\|\E_k[\ellh_k]\right\|_\infty+\left\|\E_k[b_k]\right\|_\infty\le1+2L.
\end{align*}
Since $0\le\E_k[c_k]\le1$ and $R_k$ is a sub-transition, $0\le\E_k[\Vh_k]=V^{R_k,\pi_k}(\cdot;\E_k[c_k])\le L$. Thus the two quantities in the absolute value on the first line lie in $[0,L]$. The second term is at most $L$ since $0\le P-R_k\le P$ and $q_k(x)\le u_k(x)$. By the definition of $q^\circ$ in \pref{sec:each-term},
\begin{align*}
\|q^\circ-q^\star\|_1&=\frac{1}{T}\left\|q^{P_0,\pi^{\mathrm u}}-q^\star\right\|_1\le\frac{\|q^{P_0,\pi^{\mathrm u}}\|_1+\|q^\star\|_1}{T}=\frac{2L}{T}.
\end{align*}
Recall that $\calE_0^c$ is the event that $P_0$ falls outside at least one block's confidence set defined in \pref{lem:confidence-shift}. Since $\|\qh_k-q^\circ\|_1\le2L$ and $\calG_k\setminus\calG_k^0\subseteq\calE_0^c$, the last two terms on the right-hand side of \pref{eq:app-comparator-decomposition} satisfy
\begin{align*}
\sum_{k=1}^K\E\left[\indic{\calG_k}\inner{q^\circ-q^\star,\E_k[\dhat_k]}\right]&\le \frac{2LK(1+2L)}{T},\\
\sum_{k=1}^K\E\left[\indic{\calG_k\setminus\calG_k^0}\inner{\qh_k-q^\circ,\E_k[\dhat_k]}\right]&\le2LK(1+2L)\Pr[\calE_0^c].
\end{align*}
Their sum is at most $\order(L^2/m)$ by \pref{lem:confidence-shift} and $2mK=T$.

To bound the first term on the right-hand side of \pref{eq:app-comparator-decomposition}, we define the unconstrained mirror descent update over positive state-action vectors by
\begin{align*}
z_k&\triangleq\mathop{\arg\min}_{z>0}\left\{\eta\inner{z,\dhat_k}+\Dpsi(z,\qh_k)\right\}.
\end{align*}
Its optimality condition gives $\nabla\Psi(z_k)=\nabla\Psi(\qh_k)-\eta\dhat_k$. Substituting this equality into the definition of the Bregman divergence gives
\begin{align}
\eta\inner{\qh_k-q^\circ,\dhat_k}
&=\Dpsi(q^\circ,\qh_k)-\Dpsi(q^\circ,z_k)+\Dpsi(\qh_k,z_k).\label{eq:app-omd-identity}
\end{align}
Since $16mL\eta\le1$ by the hypothesis of \pref{lem:omd-regret}, \pref{alg:main} sets $\gamma=16mL\eta$. Multiplying the bound of \pref{lem:input-size} by $\eta$ gives
\begin{align*}
\eta\qh_k(x,a)\left|\dhat_k(x,a)\right|&\le\frac{4\eta L}{\gamma}=\frac{4\eta L}{16mL\eta}=\frac{1}{4m}.
\end{align*}
For $F(y)\triangleq y\log y-\frac1m\log y$, we have $F''(y)=(y+1/m)/y^2$. At $y=\qh_k(x,a)$, both $F'(2y)-F'(y)$ and $F'(y)-F'(y/2)$ exceed $1/(4my)\ge\eta|\dhat_k(x,a)|$. The coordinate optimality condition then gives $F'(y/2)<F'(z_k(x,a))=F'(y)-\eta\dhat_k(x,a)<F'(2y)$. Since $F''>0$, $F'$ is strictly increasing, so $y/2<z_k(x,a)<2y$, and $F''\ge F''(y)/4$ throughout this interval. The curvature bound of \citet[Theorem~6, Eqs.~(4)--(6)]{bubeck2018sparsity} gives $D_F(y,z_k(x,a))\le2\eta^2\dhat_k(x,a)^2/F''(y)$.

The events $\calG_k\cap\calG_k^0$ hold on a prefix of blocks. Before the last block of this prefix, $q^\circ\in\Delta(\calP_{k+1})$, and \pref{eq:alg-update} minimizes $\Dpsi(\cdot,z_k)$ over this set. The optimality condition for the update gives
\begin{align*}
\Dpsi(q^\circ,z_k)-\Dpsi(q^\circ,\qh_{k+1})-\Dpsi(\qh_{k+1},z_k)
&=\inner{q^\circ-\qh_{k+1},\nabla\Psi(\qh_{k+1})-\nabla\Psi(z_k)}\ge0.
\end{align*}
Thus $\Dpsi(q^\circ,z_k)\ge\Dpsi(q^\circ,\qh_{k+1})$. On the last block we use $\Dpsi(q^\circ,z_k)\ge0$. Summing \pref{eq:app-omd-identity} over the prefix telescopes the comparator divergences, giving
\begin{align}
\sum_{k=1}^K\indic{\calG_k\cap\calG_k^0}\inner{\qh_k-q^\circ,\dhat_k}
&\le \frac{\Dpsi(q^\circ,\qh_1)}{\eta}
 +2\eta\sum_{k=1}^K\indic{\calG_k}\sum_{x\in\calX\setminus\{x_L\}}\sum_{a\in\calA}\frac{\qh_k(x,a)^2\dhat_k(x,a)^2}{\qh_k(x,a)+1/m}.\label{eq:app-omd-telescope}
\end{align}

For the first term on the right-hand side of \pref{eq:app-omd-telescope}, $\nabla\Psi(\qh_1)$ is constant on each layer and $q^\circ(x,a)\ge(XAT)^{-3}$. Hence $\Dpsi(q^\circ,\qh_1)=\Psi(q^\circ)-\Psi(\qh_1)\le L\log(XA)+\frac{3XA}{m}\log(XAT)\le LA\iota$.

For the second term on the right-hand side of \pref{eq:app-omd-telescope}, we fix a block with $P\in\calP_k$ and split $\dhat_k=\ellh_k+b_k$ before taking the conditional expectation. Since $(\alpha+\beta)^2\le2\alpha^2+2\beta^2$, we have
\begin{align}
&\sum_{x\in\calX\setminus\{x_L\}}\sum_{a\in\calA}\frac{\qh_k(x,a)^2\E_k[\dhat_k(x,a)^2]}{\qh_k(x,a)+1/m}\nonumber\\
&\le2\sum_{x\in\calX\setminus\{x_L\}}\sum_{a\in\calA}\frac{\qh_k(x,a)^2\E_k[\ellh_k(x,a)^2]}{\qh_k(x,a)+1/m}+2\sum_{x\in\calX\setminus\{x_L\}}\sum_{a\in\calA}\frac{\qh_k(x,a)^2\E_k[b_k(x,a)^2]}{\qh_k(x,a)+1/m}.\label{eq:app-stability-split}
\end{align}
For the first term on the right-hand side of \pref{eq:app-stability-split}, we compute the second moment of \pref{eq:loss-est}. Conditional on $\calF_{k-1}$ and $J_k$, $O_k(x,a)$ and $\qt_k(x)$ are independent. Since $0\le\ell_t(x,a)\le1$, $mO_k(x,a)$ is bounded by the number of visits to $(x,a)$ in $I_{k,2}$, which has distribution $\mathrm{Binomial}(m,q_k(x,a))$. Also, $m\qt_k(x)$ has distribution $\mathrm{Binomial}(m,q_k(x))$. Taking the product of the two conditional second moments and averaging over $J_k$ gives
\begin{align*}
\E_k[\ellh_k(x,a)^2]
&\le\frac{q_k(x,a)^2+q_k(x,a)/m}{u_k(x,a)^2}\left[\left(1+\frac{u_k(x)-q_k(x)}{u_k(x)+\gamma}\right)^2+\frac{q_k(x)(1-q_k(x))}{m(u_k(x)+\gamma)^2}\right]\\
&\le\left(1+\frac{1}{m\qh_k(x,a)}\right)\left(4+\frac{1}{mu_k(x)}\right).
\end{align*}
The last inequality uses $q_k(x,a)^2\le u_k(x,a)^2$, $q_k(x,a)\qh_k(x,a)\le u_k(x,a)^2$, and $q_k(x)\le u_k(x)$. Multiplying by the weight in \pref{eq:app-stability-split} gives
\begin{align}\label{eq:app-loss-stability}
\sum_{x\in\calX\setminus\{x_L\}}\sum_{a\in\calA}\frac{\qh_k(x,a)^2\E_k[\ellh_k(x,a)^2]}{\qh_k(x,a)+1/m}
&\le\sum_{x\in\calX\setminus\{x_L\}}\qh_k(x)\left(4+\frac{1}{mu_k(x)}\right)\le4L+X/m,
\end{align}
where the last inequality uses $\sum_{x\in\calX\setminus\{x_L\}}\qh_k(x)=L$ and $\qh_k(x)\le u_k(x)$.

For the second term on the right-hand side of \pref{eq:app-stability-split}, we compute the second moment of $b_k(x,a)$ conditional on $\calF_{k-1}$, $J_k$, and $\Vh_k$. Since $\Vh_k$ uses only $I_{k,2}$, the $m$ observations in $S_k(x,a)$ are conditionally independent. Each observation is zero unless $(x,a)$ is visited. On a visit, its mean is $[(P-R_k)\Vh_k](x,a)$ and its second moment is $\sum_{x'\in\calX_{h(x)+1}}P(x'|x,a)(\Vh_k(x')-[R_k\Vh_k](x,a))^2$. Under this conditioning, the variance of $S_k(x,a)$ is
\begin{align*}
\operatorname{Var}_k(S_k(x,a)\mid J_k,\Vh_k)
&\triangleq\E_k\left[\left(S_k(x,a)-\E_k[S_k(x,a)\mid J_k,\Vh_k]\right)^2\,\middle|\,J_k,\Vh_k\right].
\end{align*}
Substituting \pref{eq:dhat} and using that $[R_k\Vh_k](x,a)-\Vh_k(x)$ is fixed under this conditioning, we obtain
\begin{align*}
\E_k\left[b_k(x,a)^2\mid J_k,\Vh_k\right]
&=\E_k\left[\left(\frac{S_k(x,a)}{u_k(x,a)}+[R_k\Vh_k](x,a)-\Vh_k(x)\right)^2\,\middle|\,J_k,\Vh_k\right]\\
&=\left(\frac{\E_k[S_k(x,a)\mid J_k,\Vh_k]}{u_k(x,a)}+[R_k\Vh_k](x,a)-\Vh_k(x)\right)^2\\
&\hphantom{{}={}}+\frac{\operatorname{Var}_k(S_k(x,a)\mid J_k,\Vh_k)}{u_k(x,a)^2}.
\end{align*}
The conditional variance of $S_k(x,a)$ is the variance of one observation divided by $m$. Since the visit probability is $q_k(x,a)$, the moments above give
\begin{align*}
\E_k[S_k(x,a)\mid J_k,\Vh_k]&=q_k(x,a)[(P-R_k)\Vh_k](x,a),\\
\operatorname{Var}_k(S_k(x,a)\mid J_k,\Vh_k)
&=\frac{q_k(x,a)}{m}\sum_{x'\in\calX_{h(x)+1}}P(x'|x,a)\left(\Vh_k(x')-[R_k\Vh_k](x,a)\right)^2\\
&\hphantom{{}={}}-\frac{q_k(x,a)^2}{m}\left([(P-R_k)\Vh_k](x,a)\right)^2.
\end{align*}
We substitute these two expressions into the conditional second moment. Applying $(\alpha+\beta)^2\le2\alpha^2+2\beta^2$ to the squared mean and Jensen's inequality to $([(P-R_k)\Vh_k](x,a))^2$, we obtain
\begin{align}
\E_k[b_k(x,a)^2]&=\E_k\left[\E_k[b_k(x,a)^2\mid J_k,\Vh_k]\right]\nonumber\\
&\le2\E_k\left[\left([R_k\Vh_k](x,a)-\Vh_k(x)\right)^2\right]\nonumber\\
&\hphantom{{}\le{}}+\frac{2q_k(x,a)^2+q_k(x,a)/m}{u_k(x,a)^2}\sum_{x'\in\calX_{h(x)+1}}P(x'|x,a)\E_k\left[\left(\Vh_k(x')-[R_k\Vh_k](x,a)\right)^2\right].\label{eq:app-bias-second-moment}
\end{align}
In the last term on the right-hand side of \pref{eq:app-bias-second-moment}, we subtract and add $\Vh_k(x)$ inside the square. Since $q_k(x,a),\qh_k(x,a)\le u_k(x,a)$, we have
\begin{align}
\frac{q_k(x,a)^2+q_k(x,a)/m}{u_k(x,a)^2}&\le 1+\frac{1}{m\qh_k(x,a)}.\label{eq:app-bias-weight-ratio}
\end{align}
Using \pref{eq:app-bias-weight-ratio} in \pref{eq:app-bias-second-moment} and multiplying by the weight in \pref{eq:app-stability-split} gives
\begin{align}
\frac{\qh_k(x,a)^2\E_k[b_k(x,a)^2]}{\qh_k(x,a)+1/m}&\le 6\qh_k(x,a)\E_k\left[\left([R_k\Vh_k](x,a)-\Vh_k(x)\right)^2\right]\nonumber\\
&\hphantom{{}\le{}}+4\qh_k(x,a)\sum_{x'\in\calX_{h(x)+1}}P(x'|x,a)\E_k\left[\left(\Vh_k(x')-\Vh_k(x)\right)^2\right].\label{eq:app-bias-two-terms}
\end{align}
For the first term on the right-hand side of \pref{eq:app-bias-two-terms}, we use $\sum_{x'\in\calX_{h(x)+1}}R_k(x'|x,a)=1-\delta_k(x,a)$ to obtain
\begin{align}
&\left([R_k\Vh_k](x,a)-\Vh_k(x)\right)^2\nonumber\\
&=\left(\sum_{x'\in\calX_{h(x)+1}}R_k(x'|x,a)\left(\Vh_k(x')-\Vh_k(x)\right)-\delta_k(x,a)\Vh_k(x)\right)^2\nonumber\\
&\le 2\left(1-\delta_k(x,a)\right)\sum_{x'\in\calX_{h(x)+1}}R_k(x'|x,a)\left(\Vh_k(x')-\Vh_k(x)\right)^2+2\delta_k(x,a)^2\Vh_k(x)^2\nonumber\\
&\le 2\sum_{x'\in\calX_{h(x)+1}}\Ph_k(x'|x,a)\left(\Vh_k(x')-\Vh_k(x)\right)^2+2\epsilon_k(x,a)\Vh_k(x)^2,\label{eq:app-bias-lower-transition}
\end{align}
where the first inequality uses $(s-t)^2\le2s^2+2t^2$ and Cauchy--Schwarz with weights $R_k(x'|x,a)$. The second uses $1-\delta_k(x,a)\le1$, $R_k\le\Ph_k$, and $\delta_k(x,a)^2\le\epsilon_k(x,a)$ by \pref{lem:diameter}(c).

For the second term on the right-hand side of \pref{eq:app-bias-two-terms}, we first bound the squared difference for each $x'\in\calX_{h(x)+1}$. Since $\Vh_k(x')\Vh_k(x)\ge0$, we have
\begin{align}
\E_k\left[\left(\Vh_k(x')-\Vh_k(x)\right)^2\right]&\le\E_k[\Vh_k(x')^2]+\E_k[\Vh_k(x)^2]\le2\max_{z\in\calX}\E_k[\Vh_k(z)^2].\label{eq:app-bias-increment-max}
\end{align}
We next compare the average under $P$ in \pref{eq:app-bias-two-terms} with the average under $\Ph_k$ in \pref{eq:app-bias-lower-transition}. Applying $(s+t)^2\le2s^2+2t^2$ to $\sqrt P=\sqrt{\Ph_k}+(\sqrt P-\sqrt{\Ph_k})$ gives $P\le2\Ph_k+2(\sqrt P-\sqrt{\Ph_k})^2$ entrywise. Hence,
\begin{align}
&\sum_{x'\in\calX_{h(x)+1}}P(x'|x,a)\E_k\left[\left(\Vh_k(x')-\Vh_k(x)\right)^2\right]\nonumber\\
&\le2\sum_{x'\in\calX_{h(x)+1}}\Ph_k(x'|x,a)\E_k\left[\left(\Vh_k(x')-\Vh_k(x)\right)^2\right]+4\epsilon_k(x,a)\max_{z\in\calX}\E_k[\Vh_k(z)^2],\label{eq:app-bias-true-transition}
\end{align}
where we apply \pref{eq:app-bias-increment-max} to the term weighted by $(\sqrt P-\sqrt{\Ph_k})^2$ and use $\hsq(P(\cdot|x,a),\Ph_k(\cdot|x,a))\le\epsilon_k(x,a)$ from \pref{lem:diameter}(a).
We substitute the conditional expectation of \pref{eq:app-bias-lower-transition} into the first term on the right-hand side of \pref{eq:app-bias-two-terms} and \pref{eq:app-bias-true-transition} into the second. Summing over $(x,a)$ gives
\begin{align}
&\sum_{x\in\calX\setminus\{x_L\}}\sum_{a\in\calA}\frac{\qh_k(x,a)^2\E_k[b_k(x,a)^2]}{\qh_k(x,a)+1/m}\nonumber\\
&\le20\E_k\left[\sum_{x\in\calX\setminus\{x_L\}}\sum_{a\in\calA}\sum_{x'\in\calX_{h(x)+1}}\qh_k(x,a,x')\left(\Vh_k(x')-\Vh_k(x)\right)^2\right]+28v_k\max_{z\in\calX}\E_k[\Vh_k(z)^2],\label{eq:app-bias-sum}
\end{align}
where $\qh_k(x,a)\Ph_k(x'|x,a)=\qh_k(x,a,x')$ is fixed given $\calF_{k-1}$, so we are able to take $\E_k$ outside the sum. For the last term, we use $\E_k[\Vh_k(x)^2]\le\max_{z\in\calX}\E_k[\Vh_k(z)^2]$ and $\sum_{x\in\calX\setminus\{x_L\}}\sum_{a\in\calA}\qh_k(x,a)\epsilon_k(x,a)=v_k$.

For the first term on the right-hand side of \pref{eq:app-bias-sum}, we expand the square using $(v'-v)^2=(v'^2-v^2)+2v(v-v')$. For the difference of squares, summing $\qh_k(x,a,x')$ over the preceding state and action gives $\qh_k(x')$, and summing over the action and next state gives $\qh_k(x)$. Hence,
\begin{align*}
\sum_{h=0}^{L-1}\sum_{x\in\calX_h}\sum_{a\in\calA}\sum_{x'\in\calX_{h+1}}\qh_k(x,a,x')\left(\Vh_k(x')^2-\Vh_k(x)^2\right)&=\sum_{h=0}^{L-1}\left(\sum_{z\in\calX_{h+1}}\qh_k(z)\Vh_k(z)^2-\sum_{z\in\calX_h}\qh_k(z)\Vh_k(z)^2\right)\\
&=\qh_k(x_L)\Vh_k(x_L)^2-\qh_k(x_0)\Vh_k(x_0)^2=-\Vh_k(x_0)^2,
\end{align*}
where we use $\Vh_k(x_L)=0$, and $\qh_k(x_0)=1$.

For the term $2v(v-v')$, we average the next value under $\pi_k$ and $\Ph_k$ given the current state $x$. By \pref{eq:alg-c}, $c_k(x)=\sum_{a\in\calA}O_k(x,a)/(u_k(x)+\gamma)$. The Bellman recursion for $\Vh_k=V^{R_k,\pi_k}(\cdot;c_k)$ gives $\Vh_k(x)=c_k(x)+\sum_{a\in\calA}\pi_k(a|x)[R_k\Vh_k](x,a)$. Thus,
\begin{align*}
\Vh_k(x)-\sum_{a\in\calA}\pi_k(a|x)[\Ph_k\Vh_k](x,a)
&=c_k(x)-\sum_{a\in\calA}\pi_k(a|x)[(\Ph_k-R_k)\Vh_k](x,a)\le c_k(x),
\end{align*}
where the inequality uses $R_k\le\Ph_k$ and $\Vh_k\ge0$. Combining the two parts of the square expansion gives
\begin{align}
&\sum_{x\in\calX\setminus\{x_L\}}\sum_{a\in\calA}\sum_{x'\in\calX_{h(x)+1}}\qh_k(x,a,x')\left(\Vh_k(x')-\Vh_k(x)\right)^2\nonumber\\
&=-\Vh_k(x_0)^2+2\sum_{x\in\calX\setminus\{x_L\}}\qh_k(x)\Vh_k(x)\left(\Vh_k(x)-\sum_{a\in\calA}\pi_k(a|x)[\Ph_k\Vh_k](x,a)\right)\nonumber\\
&\le-\Vh_k(x_0)^2+2\sum_{x\in\calX\setminus\{x_L\}}\qh_k(x)\Vh_k(x)c_k(x)\nonumber\\
&\le 2\sqrt{\sum_{x\in\calX\setminus\{x_L\}}\qh_k(x)\Vh_k(x)^2}\sqrt{\sum_{x\in\calX\setminus\{x_L\}}\qh_k(x)c_k(x)^2}\nonumber\\
&\le 2L\sum_{x\in\calX\setminus\{x_L\}}\qh_k(x)c_k(x)^2,\label{eq:app-value-changes}
\end{align}
where the second inequality drops $-\Vh_k(x_0)^2\le0$ and applies Cauchy--Schwarz inequality. For the last inequality, $c_k\ge0$ and $R_k\le\Ph_k$ give $0\le\Vh_k\le V^{\Ph_k,\pi_k}(\cdot;c_k)$ by the Bellman recursion. Applying \pref{eq:app-value-second-moment} with $f(x,a)=c_k(x)$ then gives
\begin{align*}
\sqrt{\sum_{x\in\calX\setminus\{x_L\}}\qh_k(x)\Vh_k(x)^2}
&\le L\sqrt{\sum_{x\in\calX\setminus\{x_L\}}\qh_k(x)c_k(x)^2}.
\end{align*}

For the second term on the right-hand side of \pref{eq:app-bias-sum}, the pointwise Cauchy--Schwarz bound in the proof of \pref{eq:app-value-second-moment} gives $\Vh_k(z)^2\le L V^{\Ph_k,\pi_k}(z;c_k^2)$. Taking conditional expectations yields
\begin{align}
\max_{z\in\calX}\E_k[\Vh_k(z)^2]
&\le L\max_{z\in\calX} V^{\Ph_k,\pi_k}\left(z;\E_k[c_k^2]\right)
\le L^2\max_{x\in\calX\setminus\{x_L\}}\E_k[c_k(x)^2],\label{eq:app-block-moments}
\end{align}
where the first inequality uses that $\Ph_k$ and $\pi_k$ are fixed before block $k$, and the second uses that a trajectory from $z$ visits at most $L$ decision states. Substituting the conditional expectation of \pref{eq:app-value-changes} into the first term of \pref{eq:app-bias-sum} and \pref{eq:app-block-moments} into the second gives
\begin{align}
\sum_{x\in\calX\setminus\{x_L\}}\sum_{a\in\calA}\frac{\qh_k(x,a)^2\E_k[b_k(x,a)^2]}{\qh_k(x,a)+1/m}\le 40L\sum_{x\in\calX\setminus\{x_L\}}\qh_k(x)\E_k[c_k(x)^2]+28L^2v_k\max_{x\in\calX\setminus\{x_L\}}\E_k[c_k(x)^2].\label{eq:app-bias-cost-moments}
\end{align}

To further bound the two moments in \pref{eq:app-bias-cost-moments}, we use the definition of $c_k$ in \pref{eq:alg-c}. Conditional on $\calF_{k-1}$ and $J_k$, the number of visits to $x$ in $I_{k,2}$ has distribution $\mathrm{Binomial}(m,q_k(x))$ and is at least $m\sum_{a\in\calA}O_k(x,a)$. Hence, using $q_k(x)\le u_k(x)$, we have
\begin{align*}
\E_k[c_k(x)^2]&\le\frac{q_k(x)^2+q_k(x)/m}{(u_k(x)+\gamma)^2}
\le 1+\frac{u_k(x)}{m(u_k(x)+\gamma)^2}.
\end{align*}
Multiplying by $\qh_k(x)$ and summing over $x$ gives the first bound below. Taking the maximum over $x$ gives the second.
\begin{align}
\sum_{x\in\calX\setminus\{x_L\}}\qh_k(x)\E_k[c_k(x)^2]
&\le\sum_{x\in\calX\setminus\{x_L\}}\left(\qh_k(x)+\frac{\qh_k(x)u_k(x)}{m(u_k(x)+\gamma)^2}\right)\le L+X/m,\label{eq:app-loss-moment}\\
\max_{x\in\calX\setminus\{x_L\}}\E_k[c_k(x)^2]
&\le1+\frac1m\max_{x\in\calX\setminus\{x_L\}}\frac{u_k(x)}{(u_k(x)+\gamma)^2}\le1+\frac{1}{4m\gamma},\label{eq:app-cost-max-moment}
\end{align}
where \pref{eq:app-loss-moment} uses $\qh_k(x)\le u_k(x)$ and $\sum_{x\in\calX\setminus\{x_L\}}\qh_k(x)=L$, and \pref{eq:app-cost-max-moment} uses $(u_k(x)+\gamma)^2\ge4\gamma u_k(x)$. Substituting \pref{eq:app-loss-moment} and \pref{eq:app-cost-max-moment} into the first and second terms of \pref{eq:app-bias-cost-moments}, respectively, gives
\begin{align}\label{eq:app-bias-stability}
\sum_{x\in\calX\setminus\{x_L\}}\sum_{a\in\calA}\frac{\qh_k(x,a)^2\E_k[b_k(x,a)^2]}{\qh_k(x,a)+1/m}\le40L(L+X/m)+28L^2\left(1+\frac{1}{4m\gamma}\right)v_k.
\end{align}
Substituting \pref{eq:app-loss-stability} and \pref{eq:app-bias-stability} into the two terms on the right-hand side of \pref{eq:app-stability-split} and using $X/m\le L$, we obtain
\begin{align}
\sum_{x\in\calX\setminus\{x_L\}}\sum_{a\in\calA}\frac{\qh_k(x,a)^2\E_k[\dhat_k(x,a)^2]}{\qh_k(x,a)+1/m}
&\le\order\left(L^2+L^2\left(1+\frac{1}{m\gamma}\right)v_k\right).\label{eq:app-input-stability}
\end{align}
Since $\indic{\calG_k}$ is determined before block $k$, substituting \pref{eq:app-input-stability} into the expectation of the second term of \pref{eq:app-omd-telescope} gives
\begin{align*}
&2\eta\sum_{k=1}^K\E\left[\indic{\calG_k}\sum_{x\in\calX\setminus\{x_L\}}\sum_{a\in\calA}\frac{\qh_k(x,a)^2\E_k[\dhat_k(x,a)^2]}{\qh_k(x,a)+1/m}\right]\\
&\le\order\left(\eta KL^2+\eta L^2\left(1+\frac{1}{m\gamma}\right)W\right)\le\order\left(\eta KL^2+L^2W\right),
\end{align*}
where the last inequality uses $\eta(1+1/(m\gamma))=\eta+1/(16m^2L)\le 1/(16mL)+1/(16m^2L)\le1$, since $16mL\eta\le1$ and $\gamma=16mL\eta$. Substituting this inequality, $\Dpsi(q^\circ,\qh_1)\le LA\iota$, and the bounds on the last two terms of \pref{eq:app-comparator-decomposition} into that decomposition proves \pref{eq:bound-omd}.
\end{proof}

\section{Properties of the Confidence Sets}\label{app:aux}
In this section, we bound the distance between two transitions in a confidence set and the mass missing from its lower endpoints. We then bound the distance between the trajectory laws induced by two transitions under the same policy. We use these properties in \pref{app:analysis}.

\subsection{Confidence-Set Diameter and Lower Endpoints}

The next lemma bounds the squared Hellinger distance between transitions in $\calP_k$ and the mass missing from $R_k$. It also shows that $\Delta(\calP_k)$ contains a positive occupancy measure, as required by the log-barrier in \pref{eq:regularizer}.

\begin{lemma}\label{lem:diameter}
For every block $k$, the following hold.
\begin{enumerate}
\item[(a)] For all $\Ph,\Ph'\in\calP_k$ and every pair $(x,a)$, $\hsq\big(\Ph(\cdot|x,a),\Ph'(\cdot|x,a)\big)\le\epsilon_k(x,a)$.
\item[(b)] The set $\calP_k$ contains a transition whose entries are at least $\frac{\zeta}{X}$. In particular, $\Delta(\calP_k)$ contains a vector with positive entries.
\item[(c)] For every pair $(x,a)$ and every $\Ph\in\calP_k$, $R_k(\cdot|x,a)\le\Ph(\cdot|x,a)$ entrywise, and the missing mass $\delta_k(x,a)\triangleq1-\sum_{x'\in \calX_{h(x)+1}}R_k(x'|x,a)$ satisfies $\delta_k(x,a)\le\sqrt{\epsilon_k(x,a)}$.
\end{enumerate}
\end{lemma}

\begin{proof}
Fix a pair $(x,a)$, and let $s\triangleq\sqrt{\frac{\iota}{\max\{1,N_k(x,a)-1\}}}$, so that $\epsilon_k(x,a)=\min\{2,64Xs^2\}$.

To prove part (a), we fix $x'$ and $\Ph\in\calP_k$ and set $\bar p\triangleq\Pb_k(x'|x,a)$ and $p\triangleq\Ph(x'|x,a)$. By \pref{eq:alg-conf}, we know that $|p-\bar p|\le2s\sqrt{\bar p}+\frac{14}{3}s^2.$ We show that $|\sqrt p-\sqrt{\bar p}|\le4s$. On the one hand, $p\le\bar p+8s\sqrt{\bar p}+16s^2=(\sqrt{\bar p}+4s)^2$. On the other hand, if $\sqrt{\bar p}\le4s$, then $\sqrt p\ge\sqrt{\bar p}-4s$ trivially, and otherwise
$$(\sqrt{\bar p}-4s)^2=\bar p-8s\sqrt{\bar p}+16s^2\le\bar p-2s\sqrt{\bar p}-\frac{14}{3}s^2\le p,$$
where the first inequality uses $6s\sqrt{\bar p}\ge24s^2\ge\frac{62}{3}s^2$. Hence, two transitions in $\calP_k$ differ by at most $8s$ in every square-root entry. Summing over the at most $X$ successors gives $\hsq\le64Xs^2$, and $\hsq\le2$ always holds, which proves part (a).

To prove part (b), we choose the transition at each $(x,a)$. If $N_k(x,a)=0$, then $\Pb_k(x'|x,a)=0$ for every $x'$. The radius in \pref{eq:alg-conf} is $14\iota/3\ge1$, so we choose the uniform distribution. If $N_k(x,a)>0$, we choose $(1-\zeta)\Pb_k(\cdot|x,a)+\zeta\,\mathrm{Unif}(\calX_{h(x)+1})$. This is a distribution since $\Pb_k(\cdot|x,a)$ sums to one. Its entries differ from those of $\Pb_k(\cdot|x,a)$ by at most $\zeta\le\iota/T\le s^2$, so it satisfies \pref{eq:alg-conf}. In both cases each entry is at least $\zeta/|\calX_{h(x)+1}|\ge\zeta/X$. Every state after $x_0$ is then reached with probability at least $\zeta/X$ since the preceding layer has unit mass. Under the uniform policy every state-action pair therefore has positive occupancy.

For part (c), \pref{eq:lower-R} and \pref{eq:alg-conf} give $0\le R_k(x'|x,a)\le\Ph(x'|x,a)$ for each $x'$. If $N_k(x,a)=0$, then $M_k(x'|x,a)=0$ and hence $\Pb_k(x'|x,a)=R_k(x'|x,a)=0$ for every $x'$. Since $s^2=\iota$ and $64X\iota\ge2$, we have
\begin{align*}
\epsilon_k(x,a)&=\min\{2,64X\iota\}=2,\\
\delta_k(x,a)&=1-\sum_{x'\in\calX_{h(x)+1}}R_k(x'|x,a)=1\le\sqrt{\epsilon_k(x,a)}.
\end{align*}
If $N_k(x,a)>0$ and $64Xs^2\ge2$, then $\delta_k(x,a)\le1\le\sqrt{2}=\sqrt{\epsilon_k(x,a)}$. If $N_k(x,a)>0$ and $64Xs^2<2$, then $\sum_{x'\in\calX_{h(x)+1}}\Pb_k(x'|x,a)=1$. By \pref{eq:lower-R},
\begin{align*}
\delta_k(x,a)&=\sum_{x'\in\calX_{h(x)+1}}\left(\Pb_k(x'|x,a)-R_k(x'|x,a)\right)\\
&\le2s\sum_{x'\in\calX_{h(x)+1}}\sqrt{\Pb_k(x'|x,a)}+\frac{14}{3}|\calX_{h(x)+1}|s^2\\
&\le 2s\sqrt X+\frac{14}{3}Xs^2\le 8s\sqrt X=\sqrt{\epsilon_k(x,a)}.
\end{align*}
The second inequality uses Cauchy--Schwarz and $|\calX_{h(x)+1}|\le X$. The last inequality uses $s\sqrt X\le1$.
\end{proof}

\subsection{Comparing Trajectory Laws}
For a sub-transition $\Pt$, a policy $\pi$, and a starting state $z$, denote by $q^{\Pt,\pi}(x,a|z)$ and $q^{\Pt,\pi}(x|z)$ the conditional occupancies of the pair $(x,a)$ and the state $x$, respectively. The next lemma bounds the squared Hellinger distance between two trajectory laws by the expected sum of the squared Hellinger distances between their next-state distributions under the first transition. We use this bound in \pref{lem:reach} to compare reach probabilities under the same policy.

\begin{lemma}\label{lem:path}
Let $\Pt$ and $\Ph$ be transitions, $\pi$ a policy, and $z$ a state. The laws $\mu$ and $\nu$ of the trajectory from $z$ under $(\Pt,\pi)$ and $(\Ph,\pi)$ satisfy
$$\hsq(\mu,\nu)\le\frac{15}{2}\sum_{x\in \calX\setminus\{x_L\}}\sum_{a\in \calA}q^{\Pt,\pi}(x,a|z)\,\hsq\left(\Pt(\cdot|x,a),\Ph(\cdot|x,a)\right).$$
\end{lemma}

\begin{proof}
By the definition of the squared Hellinger distance,
\begin{align*}
\hsq(\mu,\nu)&=\sum_{\tau\text{ from }z\text{ to }x_L}\left(\sqrt{\mu(\tau)}-\sqrt{\nu(\tau)}\right)^2.
\end{align*}
For a prefix $\sigma=(z,\ldots,x)$, denote by $\mu(\sigma)$ and $\nu(\sigma)$ the probabilities under the two laws that a trajectory begins with $\sigma$. Let $f(x,a)\triangleq\hsq\left(\Pt(\cdot|x,a),\Ph(\cdot|x,a)\right)$. It suffices to prove that
\begin{align}
\sum_{\tau\text{ begins with }\sigma}\left(\sqrt{\mu(\tau)}-\sqrt{\nu(\tau)}\right)^2
&\le\frac52\left(\sqrt{\mu(\sigma)}-\sqrt{\nu(\sigma)}\right)^2+\frac{15}{2}\mu(\sigma)V^{\Pt,\pi}(x;f).\label{eq:app-stopped-expectation}
\end{align}
Every trajectory starts at $z$, and $\mu((z))=\nu((z))=1$. Thus, taking $\sigma=(z)$ in \pref{eq:app-stopped-expectation} gives
\begin{align*}
\hsq(\mu,\nu)\le\frac{15}{2}V^{\Pt,\pi}(z;f)=\frac{15}{2}\sum_{x\in\calX\setminus\{x_L\}}\sum_{a\in\calA}q^{\Pt,\pi}(x,a|z)\hsq\left(\Pt(\cdot|x,a),\Ph(\cdot|x,a)\right).
\end{align*}
The equality holds because each pair $(x,a)$ contributes $f(x,a)$ with probability $q^{\Pt,\pi}(x,a|z)$.

We prove \pref{eq:app-stopped-expectation} for every prefix from $z$ ending in $\calX_{h'}$ by backward induction from $h'=L$ to $h'=h(z)$. At $h'=L$, every such prefix $\sigma$ ends at $x_L$. The only complete trajectory beginning with $\sigma$ is $\sigma$ itself, and $V^{\Pt,\pi}(x_L;f)=0$, which gives \pref{eq:app-stopped-expectation}.

For the induction step, we fix $h'$ with $h(z)\le h'<L$ and assume that \pref{eq:app-stopped-expectation} holds for every prefix from $z$ ending in $\calX_{h'+1}$. We prove it for an arbitrary prefix $\sigma=(z,\ldots,x)$ with $x\in\calX_{h'}$. We consider two cases.

If $\nu(\sigma)\ge9\mu(\sigma)$, then $\sqrt{\nu(\sigma)}-\sqrt{\mu(\sigma)}\ge\frac23\sqrt{\nu(\sigma)}$. Summing the probabilities of all complete trajectories beginning with $\sigma$ gives
\begin{align*}
\sum_{\tau\text{ begins with }\sigma}\left(\sqrt{\mu(\tau)}-\sqrt{\nu(\tau)}\right)^2
&\le\sum_{\tau\text{ begins with }\sigma}\left(\mu(\tau)+\nu(\tau)\right)=\mu(\sigma)+\nu(\sigma)\\
&\le\frac{10}{9}\nu(\sigma)\le\frac52\left(\sqrt{\mu(\sigma)}-\sqrt{\nu(\sigma)}\right)^2.
\end{align*}
This proves \pref{eq:app-stopped-expectation} since $V^{\Pt,\pi}(x;f)\ge0$.

If $\nu(\sigma)<9\mu(\sigma)$, then $\sqrt{\mu(\sigma)\nu(\sigma)}\le3\mu(\sigma)$. We group the trajectories beginning with $\sigma$ by their next action $a\in\calA$ and state $x'\in\calX_{h'+1}$. The probabilities of observing $(\sigma,a,x')$ are
\begin{align*}
\mu(\sigma,a,x')&=\mu(\sigma)\pi(a|x)\Pt(x'|x,a),\\
\nu(\sigma,a,x')&=\nu(\sigma)\pi(a|x)\Ph(x'|x,a).
\end{align*}
Substituting these probabilities and expanding the squares gives
\begin{align}
&\sum_{a\in\calA}\sum_{x'\in\calX_{h'+1}}\left(\sqrt{\mu(\sigma,a,x')}-\sqrt{\nu(\sigma,a,x')}\right)^2-\left(\sqrt{\mu(\sigma)}-\sqrt{\nu(\sigma)}\right)^2\nonumber\\
&=\sqrt{\mu(\sigma)\nu(\sigma)}\sum_{a\in\calA}\pi(a|x)\left(2-2\sum_{x'\in\calX_{h'+1}}\sqrt{\Pt(x'|x,a)\Ph(x'|x,a)}\right)\nonumber\\
&=\sqrt{\mu(\sigma)\nu(\sigma)}\sum_{a\in\calA}\pi(a|x)f(x,a)\le3\mu(\sigma)\sum_{a\in\calA}\pi(a|x)f(x,a).\label{eq:app-prefix-increment}
\end{align}
For every $a\in\calA$ and $x'\in\calX_{h'+1}$, the prefix $(\sigma,a,x')$ ends at $x'$ in layer $h'+1$. Applying the induction hypothesis \pref{eq:app-stopped-expectation} to this prefix gives
\begin{align}
\sum_{\tau\text{ begins with }(\sigma,a,x')}\left(\sqrt{\mu(\tau)}-\sqrt{\nu(\tau)}\right)^2\le\frac52\left(\sqrt{\mu(\sigma,a,x')}-\sqrt{\nu(\sigma,a,x')}\right)^2+\frac{15}{2}\mu(\sigma,a,x')V^{\Pt,\pi}(x';f).\label{eq:app-prefix-next-layer}
\end{align}
Every complete trajectory beginning with $\sigma$ has a unique next pair $(a,x')$. Thus, summing \pref{eq:app-prefix-next-layer} over $a\in\calA$ and $x'\in\calX_{h'+1}$ gives the first inequality below.
\begin{align*}
&\sum_{\tau\text{ begins with }\sigma}\left(\sqrt{\mu(\tau)}-\sqrt{\nu(\tau)}\right)^2\\
&\le\sum_{a\in\calA}\sum_{x'\in\calX_{h'+1}}\left[\frac52\left(\sqrt{\mu(\sigma,a,x')}-\sqrt{\nu(\sigma,a,x')}\right)^2+\frac{15}{2}\mu(\sigma,a,x')V^{\Pt,\pi}(x';f)\right]\\
&\le\frac52\left(\sqrt{\mu(\sigma)}-\sqrt{\nu(\sigma)}\right)^2+\frac{15}{2}\mu(\sigma)\sum_{a\in\calA}\pi(a|x)\left(f(x,a)+\left[\Pt V^{\Pt,\pi}(\cdot;f)\right](x,a)\right).
\end{align*}
The second inequality uses \pref{eq:app-prefix-increment} and the formula for $\mu(\sigma,a,x')$. The sum over $a$ in the last line equals $V^{\Pt,\pi}(x;f)$ by the Bellman recursion. This proves \pref{eq:app-stopped-expectation} for every prefix ending in $\calX_{h'}$ and completes the induction.
\end{proof}

\section*{AI Statement}
The author developed an initial algorithmic proposal and explored its analysis through extensive discussions with GPT-5.6, GPT-6 Astra, and GPT-6.2 Astra. The author then substantially simplified and revised the algorithm and worked through and refined its analysis. The author also used Claude Opus 5.5 for language editing, literature search, reference checking, proof checking, and LaTeX editing, and used GPT-6.2 Astra and Gemini 4 Neon to proofread the draft and check it for consistency. The author determined the organization of the paper and extensively rewrote all the AI-generated materials. The author takes full responsibility for the correctness, originality, and final content of the paper.

\bibliographystyle{plainnat}
\bibliography{references}

\end{document}